\documentclass[letterpaper]{article} % DO NOT CHANGE THIS
\usepackage{aaai2026}  % DO NOT CHANGE THIS
\usepackage{times}  % DO NOT CHANGE THIS
\usepackage{helvet}  % DO NOT CHANGE THIS
\usepackage{courier}  % DO NOT CHANGE THIS
\usepackage[hyphens]{url}  % DO NOT CHANGE THIS
\usepackage{graphicx} % DO NOT CHANGE THIS
\usepackage{natbib}  % DO NOT CHANGE THIS AND DO NOT ADD ANY OPTIONS TO IT
\usepackage{caption} % DO NOT CHANGE THIS AND DO NOT ADD ANY OPTIONS TO IT
\usepackage[utf8]{inputenc}
\usepackage{textcomp}

\usepackage{graphicx} 
\usepackage{amsfonts}
\usepackage{amsmath}

\usepackage[ruled,vlined]{algorithm2e}
\usepackage{amssymb}

\usepackage{array}
\usepackage{booktabs}

\usepackage{subcaption}

\usepackage{tcolorbox}  % case study
\tcbuselibrary{breakable} % 引入 breakable 库

\usepackage{pifont}  % 红色错号
\newcommand{\cmark}{\ding{51}} % 对号
\usepackage{xcolor}

\usepackage{newfloat}
\usepackage{listings}
\DeclareCaptionStyle{ruled}{labelfont=normalfont,labelsep=colon,strut=off} % DO NOT CHANGE THIS
\title{MAO-ARAG: Multi-Agent Orchestration for Adaptive Retrieval-Augmented Generation}
\author{
    Yiqun Chen\textsuperscript{\rm 1}\equalcontrib, Erhan Zhang\textsuperscript{\rm 1}\equalcontrib, Lingyong Yan\textsuperscript{\rm 2}, Shuaiqiang Wang\textsuperscript{\rm 2}\\ 
    Jizhou Huang\textsuperscript{\rm 2}, Dawei Yin\textsuperscript{\rm 2}, Jiaxin Mao\textsuperscript{\rm 1}\thanks{Jiaxin Mao is the corresponding author.}
}
\affiliations{
    \textsuperscript{\rm 1}Renmin University of China,
    \textsuperscript{\rm 2}Baidu Inc.
    chenyiqun990321@ruc.edu.cn, erhanzhang@ruc.edu.cn, maojiaxin@gmail.com
}

\usepackage{bibentry}
\begin{document}

\maketitle

\begin{abstract}
In question-answering (QA) systems, Retrieval-Augmented Generation (RAG) has become pivotal in enhancing response accuracy and reducing hallucination issues. The architecture of RAG systems varies significantly, encompassing single-round RAG, iterative RAG, and reasoning RAG, each tailored to address different types of queries. Due to the varying complexity of real-world queries, a fixed RAG pipeline often struggles to balance performance and cost efficiency across different queries. To address this challenge, we propose an adaptive RAG framework called MAO-ARAG, which leverages multi-agent orchestration. Our adaptive RAG is conceived as a multi-turn framework. Specifically, we define multiple executor agents, representing typical RAG modules such as query reformulation agents, document selection agent, and generation agents. A planner agent intelligently selects and integrates the appropriate agents from these executors into a suitable workflow tailored for each query, striving for high-quality answers while maintaining reasonable costs. During each turn, the planner agent is trained using reinforcement learning, guided by an outcome-based reward (F1 score) and a cost-based penalty, continuously improving answer quality while keeping costs within a reasonable range. Experiments conducted on multiple QA datasets demonstrate that our approach, which dynamically plans workflows for each query, not only achieves high answer quality but also maintains both cost and latency within acceptable limits.The code of MAO-ARAG is on https://github.com/chenyiqun/Agentic-RAG.
\end{abstract}

% The architecture of RAG systems varies significantly, encompassing single-round RAG, iterative RAG, and reasoning RAG, each tailored to address different types of queries.

% Uncomment the following to link to your code, datasets, an extended version or similar.
% You must keep this block between (not within) the abstract and the main body of the paper.
% \begin{links}
%     \link{Code}{https://aaai.org/example/code}
%     \link{Datasets}{https://aaai.org/example/datasets}
%     \link{Extended version}{https://aaai.org/example/extended-version}
% \end{links}

\section{1\hspace{1em}Introduction}

Large Language Models (LLMs) have been extensively used for various tasks, including question answering \cite{asai2023self,khattab2022demonstrate}, information retrieval \cite{sun2023chatgpt, zhang2024usimagent,chen2024tourrank}, different types of reasoning \cite{huang2022towards,hao2023reasoning}, and evaluation \cite{gong2023coascore,fu2023gptscore}. Despite their wide applicability, LLMs face limitations due to their inability to update internal knowledge promptly after pre-training, making them susceptible to producing outdated or inaccurate information \cite{zhao2023survey}. To address these limitations, Retrieval-Augmented Generation (RAG) systems have been developed to boost the generative performance of LLMs by integrating relevant information from external knowledge sources, with an inherently modular architecture \cite{gao2024modular} that allows for customization to specific tasks.

% To overcome these limitations, Retrieval-Augmented Generation (RAG) has emerged to enhance the generative performance of LLMs by incorporating relevant information from external knowledge sources. 

% Existing RAG systems are inherently modular \cite{gao2024modular}, comprising various components that can be tailored to specific tasks. 

A naive RAG pipeline typically includes a retrieval model \cite{wang2022text,xiao2024c} that retrieves candidate texts, and a LLM that generates answers based on the retrieved texts. Beyond this fundamental structure, other advanced RAG pipelines incorporate additional modules such as query rewriting \cite{ma2023query,chen2025improving}, document selection \cite{ke2024bridging,li2024rag}, and self-reflection \cite{asai2023self}. Recently, methods like Search-o1 \cite{li2025search} and Search-r1 \cite{jin2025search} have further advanced the capabilities of RAG systems by integrating reasoning processes.

\begin{figure}[t]
  \centering
  \includegraphics[width=0.475\textwidth]{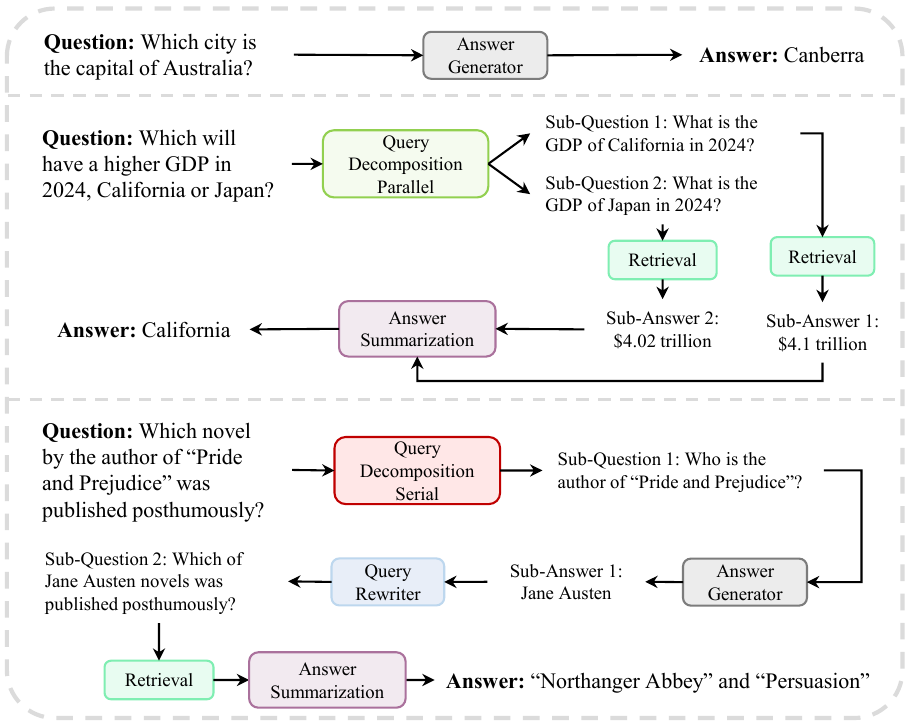}
  % \vspace{-6mm}
  \caption{The appropriate workflows for different types of queries are highly heterogeneous.}
  \label{intro}
\end{figure}

These RAG systems, with their diverse components, are suited to different scenarios. Naive RAG are ideal for straightforward queries, offering benefits such as lower costs and reduced latency. However, they tend to struggle with more complex queries. On the other hand, more sophisticated RAG systems excel in handling intricate questions but at the cost of increased LLM usage and higher latency. In real-world QA systems, queries often vary widely in type and difficulty. And as shown in Figure \ref{intro}, the ideal workflows vary across different types of queries. Therefore, it is challenging for a fixed RAG pipeline to consistently deliver high-quality answers across diverse queries while keeping costs—such as LLM token usage and system latency—within a reasonable range.

To address this challenge, we propose an \textbf{A}daptive multi-turn \textbf{RAG} framework called MAO-ARAG, utilizing \textbf{M}ulti-\textbf{A}gent \textbf{O}rchestration. Within this framework, we define multiple executor agents comprising common modules found in existing RAG systems, such as query reformulation, retriever, document selection, and answer generation, etc. At the core of this framework lies a planner agent that selects suitable executors for each query and orchestrates them to form a query-specific workflow. To improve the effectiveness and efficiency of the planner agent's orchestration, we adopt the Proximal Policy Optimization (PPO) algorithm~\cite{schulman2017proximal}, guided by an outcome-based reward (F1 score) and a cost-based penalty. This approach ensures that the constructed pipeline achieves high answer quality while keeping costs, such as resource consumption and latency, within reasonable bounds.

Our main contributions are as follows:

\begin{itemize}
\item We propose MAO-ARAG, a novel multi-agent framework for adaptive RAG that features a planner agent to dynamically select and compose multiple executor agents—modular components commonly used in RAG systems—into a query-specific workflow.
\item We propose a PPO-based training algorithm that incorporates outcome-based rewards and cost-based penalties to improve the planner agent’s ability to balance answer quality and computational cost.
\item We conduct extensive experiments on multiple QA benchmarks to validate that the proposed MAO-ARAG framework can tailor a suitable RAG pipeline for each query, achieving high answer quality and maintaining appropriate cost.
\end{itemize}

\section{2\hspace{1em}Related Work}

\subsection{2.1\hspace{0.5em}Different Modules in RAG System}

\textbf{Retrieval Model} plays a fundamental role in the RAG system, which supplies external knowledge to the LLM-based generator to generate final answers. Within the context of retrieval models for RAG, BM25 \cite{robertson1994some} stands out as a traditional yet effective sparse retrieval model. In contrast, Contriever \cite{izacard2021unsupervised}, BGE \cite{wang2022text}, and E5 \cite{xiao2024c} are designed to produce dense embeddings, making them effective dense retrieval models. Lastly, ColBERT \cite{khattab2020colbert} improves information retrieval accuracy and efficiency by employing multi-vector representations and a “late interaction” mechanism.

% \textbf{Reranking Model} has become essential in enhancing the relevance and accuracy of search results in RAG systems. MiniLM \cite{wang2020minilm,wang2020minilmv2} is a lightweight, distilled version of BERT, designed to efficiently rerank documents by leveraging semantic similarity while maintaining high performance and computational efficiency. Meanwhile, the BGE-reranker \cite{xiao2024c} is also an advanced reranking model that incorporates sophisticated neural network architectures and fine-tuning techniques to capture complex query-document relationships and achieve superior performance in various reranking tasks.

\textbf{Query Reformulation} is to rewrite or decompose initial query in RAG, and are introduced in RRR \cite{ma2023query} and DMQR-RAG \cite{li2024dmqr}.

\textbf{Document Selection} is to select helpful information from the noise candidate documents. BGM \cite{ke2024bridging} and RAG-DDR \cite{li2024rag} both utilize this module.

\textbf{Answer Generator} is responsible to output the answer to the input query. There are many open-source LLMs, such as Deepseek \cite{bi2024deepseek}, Llama \cite{grattafiori2024llama}, Qwen \cite{yang2025qwen3}, and many close-source LLMs, such as GPT \cite{brown2020language} and Gemini \cite{team2023gemini}, which can be considered as an answer generator.

\subsection{2.2\hspace{0.5em}Typical Workflows in RAG System}

\textbf{Single-Round RAG} \quad The modules in single-turn RAG are organized in a linear way. RRR \cite{ma2023query} propose a Rewrite-Retrieve-Read framework and BGM \cite{ke2024bridging} introduce a selection-generation paradigm. In addition, RAG-DDR \cite{li2024rag} and MMOA-RAG \cite{chen2025improving} also propose a linear RAG pipeline. All these method utilized reinforcement learning algorithm to optimize single or multiple modules in RAG pipelines.

\textbf{Iterative RAG} \quad The RAG pipeline in iterative RAG is a loop structure. ITER-RETGEN \cite{shao2023enhancing} is a method that improves retrieval-augmented large language models by iteratively integrating retrieval and generation processes. SELF-RAG \cite{asai2023self} boosts the quality and factual accuracy of language models through a process of self-reflective retrieval and generation. DRAGIN \cite{su2024dragin} is a framework that dynamically addresses the real-time information needs of large language models during text generation, enhancing their performance on tasks that require extensive knowledge. SMARTRAG \cite{gao2024smartrag} utilizes PPO to optimize an iterative RAG framework with answer-based reward.

\textbf{Reasoning RAG} \quad Search-o1 \cite{li2025search} enhances the RAG utilizing reasoning ability of LLM. After Deepseek-r1 \cite{guo2025deepseek}, some works introduce training the reasoning LLM to improve the performance in RAG. For example, Search-r1 \cite{jin2025search} and R1-Searcher \cite{song2025r1} both use answer-based reward to improve the reasoning in RAG.

\section{3\hspace{1em}Methods}

\begin{figure*}[t]
  \centering
  \includegraphics[width=1.0 \textwidth]{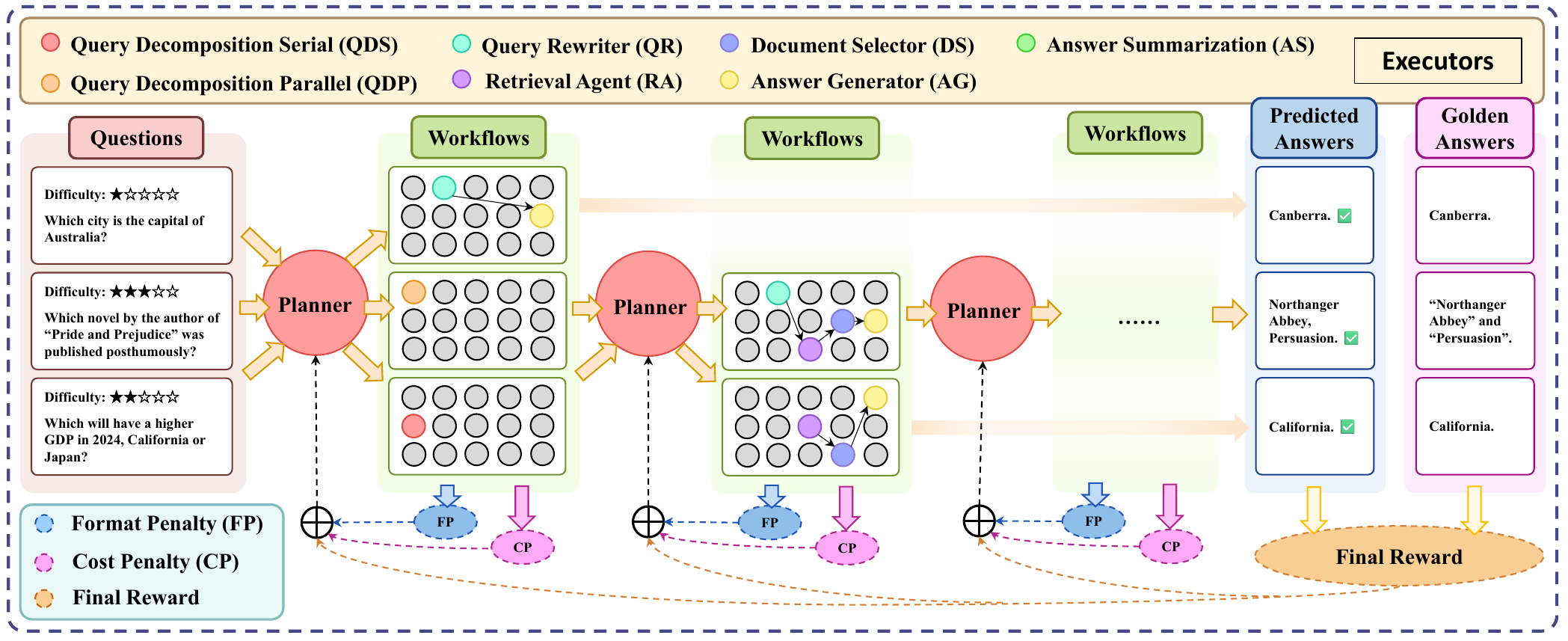}
  \caption{The overall framework of MAO-ARAG.}
  \label{framework}
\end{figure*}

\subsection{3.1\hspace{0.5em} Modeling RAG as a Multiagent Semi-Markov Decision Process}

% We model our MAO-ARAG as a Multiagent Semi-Markov Decision Process (MSMDP) \cite{ghavamzadeh2006hierarchical}, which is well-suited for modeling systems involving multiple agents with distinct roles. 

To capture the requirements of orchestrating heterogeneous agents across varying scenarios, we model the RAG system as a Multiagent Semi-Markov Decision Process (MSMDP)~\cite{ghavamzadeh2006hierarchical}, which effectively captures coordination among agents with distinct roles.

% Specifically, our approach involves a planner agent and several executor agents, each contributing to the decision-making process in a coordinated manner.

An MSMDP extends the traditional Markov Decision Process (MDP) \cite{sutton1998reinforcement} by accommodating multiple agents and allowing for actions of variable durations. Formally, an MSMDP can be defined as a tuple $\langle S, A, P, R, \gamma, T \rangle$. $S$ is the state space. $A = \{A_1, A_2, \ldots, A_n\}$ is a set of action sets, where $A_i$ is the set of actions available to agent $i$. $P: S \times A \times S \rightarrow [0, 1]$ is the state transition probability function. $R: S \times A \rightarrow \mathbb{R}$ is the reward function, providing feedback to the agents based on the current state and actions taken. $T: S \times A \rightarrow \mathbb{R}^+$ is a function representing the duration of executing an action.

To implement an adaptive RAG, we introduce MAO-ARAG, which employs a planner to coordinate multiple executors\footnote{The prompt for each agent is detailed in Appendix C.}, as illustrated in Figure \ref{framework}. 

\textbf{The Planner} is responsible for designing an appropriate workflow for a given question or a rewritten sub-question, and the workflow is composed of a subset of the executors. 

\textbf{The executors} encompass several commonly used modules in the current modular RAG process, mainly including:

\begin{itemize}
    \item \textbf{Query Decomposition Serial (QDS)}: This module serially decomposes a given question into several sub-questions that have sequential dependencies. The answer to a later sub-question often depends on the answers to preceding ones.
    
    \item \textbf{Query Decomposition Parallel (QDP)}: This module decomposes a given question into multiple independent sub-questions that can be searched in parallel.
    
    \item \textbf{Query Rewriter (QR)}: This module rewrites a question into a clearer and more searchable version.
    
    \item \textbf{Document Selector (DS)}: Given a question and multiple candidate documents, this module selects documents that are helpful for answering the question and excludes those that are not.
    
    \item \textbf{Retrieval Agent (RA)}: This is a search engine that takes a question as input and returns the top $k$ most relevant candidate documents from a corpus.
    
    \item \textbf{Answer Generator (AG)}: This module generates an answer to a given question, which may be informed by reference documents or generated independently.
    
    \item \textbf{Answer Summarization (AS)}: Based on the sub-questions and their respective sub-answers, this module provides an answer to the initial question.
\end{itemize}

% In our framework, the planner agent is responsible for high-level decision-making and planning dynamic workflows for orchestrating the executor agents. The executor agents then carry out specific tasks according to the workflows given by the planner agent. 

In MSMDP, since the duration of an action $T$ is not fixed, the MSMDP allows for effective coordination and optimization across different time scales and agent roles, making it a suitable way to model an adaptive RAG. By leveraging the MSMDP framework, our MAO-ARAG efficiently integrates the decision-making processes of the planner agent and executor agents.

% \subsection{3.2\hspace{0.5em}Overall of Our Method}

\subsection{3.2\hspace{0.5em}Essential Elements of RL}

In our framework, the workflow plays a crucial role in determining the final answers, and since the planner is responsible for generating this workflow, optimizing the planner becomes essential and important. The framework involves multiple rounds during the whole rollout process, with each round requiring the planner to design a appropriate workflow for the given (sub-)question. Moreover, our optimization goals are not limited to enhancing the answer qualities; they also include reducing cost and latency, making this a multi-objective optimization problem. Taking these factors into account, we employ a reinforcement learning approach (PPO algorithm) to optimize the parameters of planner agent with an outcome-based reward and cost-based penalty terms. 

In the following, we will introduce the essential elements of planner, which mainly contains Observation, Action Space and Reward Function:

\begin{itemize}

\item \textbf{Observation} of planner is defined as Equation (\ref{observation}), which contains the prompt of planner $Prompt_{planner}$ and a given question $q$. And $q$ is a initial question or a sub-question.

\begin{equation}\small
O_{planner} = \left\{ Prompt_{planner}, q \right\}
\label{observation}
\end{equation}

\item\textbf{Action Space} of planner is the abbreviation of each executors, which is defined as Equation (\ref{action space}). The output of planner is a combination of the abbreviations of executors in the action space.

\begin{equation}
A_{planner} = \left\{ \textbf{QDS, QDP, QR, DS, RA, AG, AS} \right\}
\label{action space}
\end{equation}

% \item\textbf{Reward Function} of the planner consists of three components. As shown in Equation (\ref{F1 score}), the first component is the F1 score between the answer $a_{predicted}$ predicted by the RAG and the golden answer $a_{golden}$, which is one of the performance metrics we aim to optimize and is the Final Reward in Figure \ref{framework}.

\item \textbf{Reward Function} of the planner comprises three components. The first component, shown in Equation (\ref{F1 score}), is the F1 score calculated between the predicted answer, $a_{\text{predicted}}$, generated by the RAG, and the golden answer, $a_{\text{golden}}$. This F1 score serves as one of the key performance metrics we strive to optimize and is also the Final Reward illustrated in Figure \ref{framework}.

\begin{equation}\small
R_{f1} = F1(a_{\text{predicted}}, a_{\text{golden}})
\label{F1 score}
\end{equation}

The second component is a penalty term concerning the cost, denoted as Cost Penalty (CP) in Figure \ref{framework}, which is defined as Equation (\ref{CP}). 

% \begin{equation}
% R_{CP} = Token_{cost} + \alpha \cdot Turn_{cost} + \beta \cdot \mathbb{I}(S)
% \label{CP}
% \end{equation}

\begin{equation}\small
R_{CP} = Token_{\text{cost}} + Turn_{\text{cost}} + \mathbb{I}(S)
\label{CP}
\end{equation}

In Equation (\ref{CP}), $Token_{\text{cost}}$ represents the token cost of the workflow provided by the planner, scaled to a range between 0 and 1. Similarly, $Turn_{\text{cost}}$ denotes the cost associated with latency. If a given workflow brings more turns later, the $Turn_{\text{cost}}$ will be larger. But $Turn_{\text{cost}}$ is also scaled between 0 to 1. The function $\mathbb{I}(S)$ is an indicator for the search engine call. If the Retrieval Agent (RA) is utilized in the workflow, then $\mathbb{I}(S) = 1$; otherwise, $\mathbb{I}(S) = 0$. As for the specifics of \textbf{how and why these three cost penalties are scaled, you can refer to Appendix A}.
% \footnote{As for the specifics of how and why these three cost penalties are scaled, you can refer to Appendix A.}

% \ref{How and why should the cost-based penalty terms be scaled?}

The third component concerns the penalty term related to the workflow format, denoted as Format Penalty (FP) in Figure \ref{framework}, which is defined as Equation (\ref{FP}). 

\begin{equation}\small
R_{FP} = \mathbb{I}(workflow)
\label{FP}
\end{equation}

In Equation (\ref{FP}), only when the workflow format is correct and executable is $R_{FP}$ equal to 0; otherwise, it is 1.

Finally, the total reward of the planner can be defined as Equation (\ref{reward fuction}), which consists of $R_{f1}$, $R_{CP}$, and $R_{FP}$. $\alpha$ is a hyperparameter.

\begin{equation}\small
R_{planner} = R_{f1} - \alpha \cdot R_{CP} - R_{FP}
\label{reward fuction}
\end{equation}

\end{itemize}

\subsection{3.3\hspace{0.5em}Traning Process of RL}

\begin{algorithm}[!h] \small
  \DontPrintSemicolon
  \SetKwInOut{KwOutput}{Output} % Define KwOutput before using
  \textbf{Initialize}: The parameters of the Actor model $\theta$, the Critic model $\phi$, the initial model $\theta_{\text{init}}$, and a replay buffer $\mathcal{M}=\varnothing$.\\
  \textbf{Inputs}: Dataset with initial questions $q$ and corresponding golden answers $Ans_{\text{golden}}$\\
  % \For{$epoch \gets 1$ \KwTo $N\_epoch$}{
    \For{$batch \gets 1$ \KwTo $N\_batch$}{
        \CommentSty{\text{// Collect Data}} \\
        \For{each question $q_{init} \in batch$}{
            \CommentSty{\text{// Rollout Multiple Turns}} \\
            \For{$turn_i$ in MAX\_TURN}{
                \CommentSty{\text{// Planner}} \\
                Determine the given question $q$ to planner. ($q$ is $q_{init}$ or its sub-question $q_{sub}$.)\;
                Construct observation $O_{planner}^i$ according to Equation (\ref{observation}).\;
                Get the workflow $w$ to the given question.\;
                Get the Format Penalty (FP) $R_{FP}^i$ for this $turn_i$.\;

                \CommentSty{\text{// Execute the workflow}} \\
                Following the workflow $w$, execute each executor involved.\;
                Get the Cost Penalty (CP) $R_{CP}^i$ for this $turn_i$.\;

                Store tuple $\mathcal{T}_i = (O_{planner}^{i}, w_i, R_{FP}^i, R_{CP}^i)$ in the replay buffer $\mathcal{M}$
            }
            Get the predicted answer $a_{\text{predicted}}$ for $q_{init}$.\;
            Compute the F1 score as $R_{f1}$ for $q_{init}$.\;
            \CommentSty{\text{// Update the data}} \\
            \For{each $turn_i$ in MAX\_TURN}{
                Calculate the total reward $R_{planner}^i$ for $turn_i$ according to Equation (\ref{reward fuction}).\;
                Update the tuple $\mathcal{T}_i = (O_{planner}^{i}, w_i, R_{planner}^i)$ in the replay buffer $\mathcal{M}$.
            }
          }
        
        \CommentSty{\text{// Policy and Value Optimization}} \\
        \For{each question $q \in batch$}{
            Compute the advantage function $\hat{A}_{{\pi}_{\theta}}^{t}$ using GAE\;
            Calculate the loss of the Actor $\mathcal L_{\text{Actor}}(\theta)$ and Critic model $\mathcal L_{\text{Critic}}(\phi)$\;
            Update the parameters of models through the overall loss function $\mathcal L(\theta,\phi)$ in Equation (\ref{ppo_loss})
        }
        Clear the replay buffer $\mathcal{M}$ to $\varnothing$\;
    }
  % }
  \KwOutput{A well-trained planner: Actor model with parameters $\theta_{\text{trained}}$}
  \caption{The Training Process of MAO-ARAG}
\end{algorithm}

As illustrated in Figure \ref{framework}, we model the entire rollout as a multi-turn process and utilize PPO algorithm to optimize the planner to get better evaluation metrics. The training process of MAO-ARAG is shown in \textbf{Algorithm 1}. The parameters for the actor model and the critic model are denoted as $\theta$ and $\phi$, respectively, and the reference model is denoted as $\theta_{init}$. In each turn, a planner is responsible for designing an adaptive workflow $w$ based on either the initial problem $q_{init}$ or its reformulated sub-question $q_{sub}$. Subsequently, the executors implement this workflow. Upon the completion of all turns, a predicted answer $a_{\text{prediected}}$ is obtained, which is then evaluated against the golden answer $a_{\text{golden}}$ using the F1 score. This F1 score serves as a shared reward across all turns. Additionally, each turn incorporates a Format Penalty (FP) and a Cost Penalty (CP) to make the workflow executable and balance the overall cost. Following this, we employ the PPO algorithm to update the planner's parameters based on the data collected in each turn. The overall loss function of PPO, $\mathcal{L}(\theta, \phi)$, consists of two terms: $\mathcal{L}_{\text{Actor}}(\theta)$ and $\mathcal{L}_{\text{Critic}}(\phi)$:

\begin{equation}\small
\mathcal L(\theta,\phi) = \mathcal L_{\text{Actor}}(\theta) + \mathcal L_{\text{Critic}}(\phi)
\label{ppo_loss}
\end{equation}

% The actor loss $\mathcal{L}_{\text{Actor}}(\theta)$ can be defined as Equation (\ref{actor_loss}). The term $r_t$ in Equation (\ref{ratio}) denotes the importance sampling ratio, which measures the difference between the new and old policies. The expression $\hat{A}_{\pi_{\theta}}^{t}$ in Equation (\ref{advantage}) is the advantage function, estimated using Generalized Advantage Estimation (GAE) \cite{schulman2015high}. The variable $\delta_t$ in Equation (\ref{td error}) is known as the temporal difference (TD) error at time step $t$.

The actor loss $\mathcal{L}_{\text{Actor}}(\theta)$ can be defined as Equation (\ref{actor_loss}). The term $r_{t} = \frac{\pi_\theta(a_t \mid s_t)}{\pi_{\theta_{old}}(a_t \mid s_t)}$ denotes the importance sampling ratio, which measures the difference between the new and old policies. The expression $\hat{A}_{{\pi}_{\theta}}^{t} = \sum_{l=0}^{\infty} (\gamma\lambda)^l \delta_{t+l}$ is the advantage function, estimated using GAE \cite{schulman2015high}. The variable $\delta_t = R(s_t, a_t) + \gamma V_{\phi}(s_{t+1}) - V_{\phi}(s_t)$ is known as the temporal difference (TD) error at time step $t$.

\begin{equation}\small
\mathcal{L_{\text{Actor}}}(\theta) = \sum_t \min \left( r_{t} \hat{A}_{{\pi}_{\theta}}^{t}, \ \text{clip} \left( r_{t}, 1-\epsilon, 1+\epsilon \right) \hat{A}_{{\pi}_{\theta}}^{t} \right)
\label{actor_loss}
\end{equation}

% \begin{equation}\small
% r_{t} = \frac{\pi_\theta(a_t \mid s_t)}{\pi_{\theta_{old}}(a_t \mid s_t)}
% \label{ratio}
% \end{equation}

% \begin{equation}\small
% \hat{A}_{{\pi}_{\theta}}^{t} = \sum_{l=0}^{\infty} (\gamma\lambda)^l \delta_{t+l}
% \label{advantage}
% \end{equation}

% \begin{equation}\small
% \delta_t = R(s_t, a_t) + \gamma V_{\phi}(s_{t+1}) - V_{\phi}(s_t)
% \label{td error}
% \end{equation}

The Equation (\ref{final_reward}) is similar with the reward in PPO training for LLM \cite{ouyang2022training}. $R_{planner}$ contains three components defined in Equation (\ref{reward fuction}). 

\begin{equation}\small
R(s_t, a_t) = 
\begin{cases} 
0, & \text{if } t < T \\
R_{planner} - \beta \cdot \log\left(\frac{\pi_{\theta}(w \mid O_{planner})}{\pi_{\theta_{init}}(w \mid O_{planner})}\right), & \text{if } t = T
\end{cases}
\label{final_reward}
\end{equation}

The critic loss $\mathcal{L}_{\text{Critic}}(\phi)$ is defined in Equation (\ref{value_loss}), employing a clipping operation similar to the actor loss. Here, $\Delta V_{t} = V_{\phi}^{t} - V_{\text{target}}^{t}$, where $V_{\phi}^{t} = V_{\phi}(s_t)$. The term $V_{\text{target}}^{t}$ represents the cumulative return and $s_t$ is the state-values.

\begin{equation}\small
\mathcal{L}_{\text{Critic}}(\phi) = \sum_t \max \left[ (\Delta V_{t})^2, \left( \text{clip}\left(V_{\phi}^{t}, V_{\phi_{\text{old}}}^{t} \pm \epsilon\right) - V_{\text{target}}^{t} \right)^2 \right]
\label{value_loss}
\end{equation}

After multiple steps of training, we can obtain a well-trained planner agent that can customize an appropriate workflow for each query.

\section{4\hspace{1em}Experiments}

Our experiments mainly focus on the following research questions:

\begin{itemize}
\item \textbf{RQ.1}: Can MAO-ARAG outperform the existing common RAG pipeline?
\item \textbf{RQ.2}: Is MAO-ARAG an efficient method? In other words, can MAO-ARAG achieve good performance while keeping costs within a reasonable range?
\item \textbf{RQ.3}: How does $\alpha$ in Equation (\ref{reward fuction}) affect the learned strategies?
\item \textbf{RQ.4}: Can we use different LLMs as the planner and executors in MAO-ARAG? For example, can we use a smaller planner for efficiency? Can we leverage alternative LLMs to support the executors for different trade-offs in effectiveness and cost?
\end{itemize}

\subsection{4.1\hspace{0.5em}Experimental Setup}

\textbf{Datasets} \quad To evaluate the effectiveness of our MAO-ARAG, we conduct experiments on a diverse set of open-domain question answering (QA) benchmarks:

\begin{itemize}
    \item \textbf{Single-hop QA}: We include Natural Questions (NQ) \cite{kwiatkowski2019natural}, PopQA \cite{mallen2022not}, and AmbigQA \cite{min2020ambigqa}.
    \item \textbf{Multi-hop QA}: We also use HotpotQA \cite{yang2018hotpotqa}, 2WikiMultiHopQA \cite{ho2020constructing}, Musique \cite{trivedi2022musique}, and Bamboogle \cite{press2022measuring} to test the ability of different methods.
\end{itemize}

\textbf{Corpus and Retriever} \quad For all retrieval-based methods, we utilize Wikipedia as the corpus \cite{karpukhin2020dense}. Retriever is performed using E5 \cite{wang2022text}.

\textbf{Evaluation Metrics} \quad We evaluate model performance using F1 score. We also utilize the token cost, retriever call times, and turn number as the cost metrics.

\textbf{Models} \quad we mainly employ \textbf{Qwen2.5-7B-Instruct} \cite{team2024qwen2} as the planner, responsible for analyzing the input query and generating an appropriate workflow. For the executor agents, we utilize \textbf{GPT-4o-Mini} \cite{hurst2024gpt} as the backbone to perform the corresponding functions.

\textbf{Baselines} \quad We compare our approach with different types of baselines\footnote{For a fair comparison, we re-implemented all the baselines based on GPT-4o-mini, except Search-r1. Since Search-r1 can be considered as the RL training version of Search-o1, their workflows are essentially very similar. Therefore, we adopted the setup from the original Search-r1 paper, and re-implemented it based on Qwen2.5-7B-Instruct model and RL reasoning training.}:

\begin{itemize}
    \item \textbf{Singel-Round RAG}: (1) \textbf{LLM w/o RAG}: Answers are generated solely based on the internal knowledge of the LLM. (2) \textbf{Vanilla RAG}: A conventional RAG setup where retrieved documents are used to generate answers. (3) \textbf{RRR} \cite{ma2023query}: Introduce query reformulation in RAG. (4) \textbf{BGM} \cite{ke2024bridging}: Add a documents selection module in RAG pipeline. (5) \textbf{MMOA-RAG} \cite{chen2025improving}: The workflow contains query rewriter, retriever, document selector, answer generator.
    \item \textbf{Iterative RAG}: (6) \textbf{Self-RAG} \cite{asai2310learning}: Combines adaptive retrieval with self-reflection to enhance answer reliability and precision.
    \item \textbf{Agentic RAG}: (7) \textbf{Search-o1} \cite{li2025search}: Incorporate an agentic retrieval mechanism and a dynamic workflow. (8) \textbf{Search-r1} \cite{jin2025search}: Utilize RL reasoning training to enhance the agentic RAG.
\end{itemize}

% \begin{itemize}
%     \item \textbf{LLM w/o RAG}: Answers are generated solely based on the internal knowledge of the LLM without access to external documents.
%     \item \textbf{Vanilla RAG}: A conventional RAG setup where retrieved documents are directly appended to the query.
%     \item \textbf{RRR} \cite{ma2023query}: Reinforcement-based Query Rewriting; a lightweight query rewriting model trained to improve retrieval quality via reward signals tied to answer generation.
%     \item \textbf{BGM} \cite{ke2024bridging}: A bridge module is trained using PPO to filter and rank documents that are more likely to be helpful, enhancing retrieval precision.
%     \item \textbf{MMOA-RAG} \cite{chen2025improving}: Models the modular RAG pipeline as a multi-agent RL problem, jointly optimizing retrieval, filtering, and answer generation for aligned objectives.
%     \item \textbf{Self-RAG} \cite{asai2310learning}: Combines adaptive retrieval with self-reflection to enhance answer reliability and precision.
%     \item \textbf{Search-r1} \cite{jin2025search}: Extends RL-based reasoning by enabling LLMs to autonomously generate multiple search queries during multi-step reasoning. 
%     % It leverages token-masking and a simplified reward to stabilize training.
%     \item \textbf{Search-o1} \cite{li2025search}: Incorporates an agentic retrieval mechanism along with a Reason-in-Documents module to refine noisy documents before reasoning.
% \end{itemize}

\begin{table*}[t] \scriptsize
\caption{F1 performance (\%) of various methods across datasets. The font with the highest score in each dataset is bold, and the second highest score is underlined. $\Delta$ indicates the improvement of MAO-ARAG over the best baseline.}
\centering
\renewcommand{\arraystretch}{0.8}
\begin{tabular}{>{\centering\arraybackslash}m{4.0cm}
                >{\centering\arraybackslash}m{1.2cm}
                >{\centering\arraybackslash}m{1.2cm}
                >{\centering\arraybackslash}m{1.2cm}
                >{\centering\arraybackslash}m{1.2cm}
                >{\centering\arraybackslash}m{1.2cm}
                >{\centering\arraybackslash}m{1.2cm}
                >{\centering\arraybackslash}m{1.2cm}
                >{\centering\arraybackslash}m{1.2cm}}
\toprule
\textbf{Methods} & \textbf{NQ} & \textbf{PopQA} & \textbf{AmbigQA} & \textbf{HotpotQA} & \textbf{2Wiki} & \textbf{Musique} & \textbf{Bamboogle} & \textbf{Average} \\
\midrule
LLM w/o RAG & 39.96 & 30.99 & 49.90 & 42.38 & 33.49 & 20.74 & 36.15 & 36.23 \\
Vanilla RAG & 48.02 & 44.23 & \underline{59.04} & 49.54 & 37.62 & 25.66 & 43.45 & 43.94 \\
RRR \cite{ma2023query} & 46.27 & 41.59 & 56.15 & 43.14 & 29.77 & 23.21 & 37.81 & 39.71 \\
BGM \cite{ke2024bridging} & 48.41 & \underline{45.39} & \textbf{59.25} & 49.58 & 36.79 & 25.60 & 44.10 & 44.16 \\
MMOA-RAG \cite{chen2025improving} & 46.88 & 40.26 & 55.88 & 43.19 & 30.40 & 21.78 & 36.53 & 39.28 \\
Self-RAG \cite{asai2023self} & 41.60 & 34.25 & 52.06 & 47.94 & 39.53 & 32.88 & 56.33 & 43.51 \\
Search-r1 \cite{jin2025search} & 42.22 & 43.35 & 52.50 & 44.44 & 34.13 & 21.44 & 37.83 & 39.42 \\
Search-o1 \cite{li2025search} & 46.68 & 43.12 & 56.93 & \underline{53.75} & \underline{47.26} & \textbf{39.51} & \underline{61.58} & \underline{49.83} \\
\midrule
MAO-ARAG w/o train & \underline{50.57} & 32.73 & 55.15 & 49.68 & 40.75 & 32.36 & 49.41 & 44.38 \\
MAO-ARAG & \textbf{54.50} & \textbf{54.16} & 57.80 & \textbf{53.80} & \textbf{47.69} & \underline{37.33} & \textbf{65.09} & \textbf{52.91} \\
$\Delta$ & \textbf{+6.09} & \textbf{+8.77} & -1.45 & \textbf{+0.05} & \textbf{+0.43} & -2.18 & \textbf{+3.51} & \textbf{+3.08} \\
\bottomrule
\end{tabular}
\label{exp_f1_across_datasets}
\end{table*}

\subsection{4.2\hspace{0.5em}Performance of Different Methods (RQ.1)}

We evaluated various algorithms across the seven datasets presented in Table \ref{exp_f1_across_datasets}, focusing on their F1 scores for comparison. Our training utilized only 2400 question-answer pairs from the NQ training dataset and 4800 pairs from the HotpotQA training dataset, followed by testing on all seven datasets. To reduce testing costs, we randomly selected 1000 question-answer pairs from the official test sets of each dataset (with Bamboogle having only 125 pairs).

In Table \ref{exp_f1_across_datasets}, “MAO-ARAG w/o train” signifies the use of the untrained Qwen2.5-7B-Instruct as the planner. Meanwhile, “MAO-RAG” represents our method, trained via RL training, where the $\alpha$ hyperparameter in Equation \ref{reward fuction} is set to zero, indicating an exclusive focus on optimizing the F1 score without considering cost reduction.

Table \ref{exp_f1_across_datasets} reveals that the average F1 score of MAO-ARAG w/o train across the seven datasets ranks just below Search-o1 among the baselines, suggesting that even an untrained planner can effectively organize and manage executors. Moreover, our MAO-ARAG method achieved the highest performance on 5 out of the 7 datasets, with an average F1 score of 52.91. This is 3.08 points higher than the best baseline, Search-o1, which scored 49.83, and 8.53 points higher than MAO-ARAG w/o train, which had a score of 44.38. These results highlight the effectiveness of our optimization mechanism for the planner in multi-turn adaptive RAG, demonstrating its capability to effectively select and arrange executors to achieve the goal of optimizing the F1 score.\footnote{The case study can be seen in Appendix E.}

\subsection{4.3\hspace{0.5em}Cost-Performance Trade-Off (RQ.2)}

\begin{figure*}[h]
\centering
\begin{subfigure}{0.33\textwidth}
  \centering
  \includegraphics[width=\linewidth]{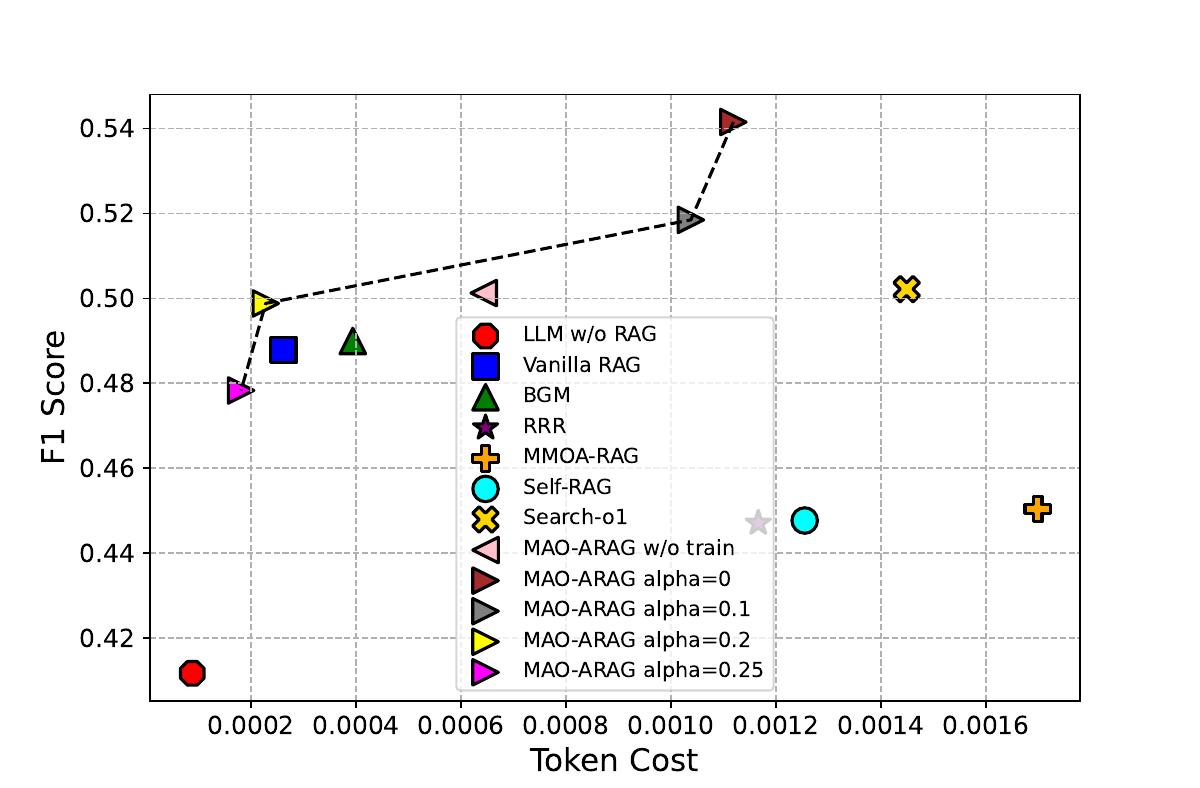}
  \caption{F1 score vs. Token Cost}
  \label{f1_cost}
\end{subfigure}
\hfill
\begin{subfigure}{0.33\textwidth}
  \centering
  \includegraphics[width=\linewidth]{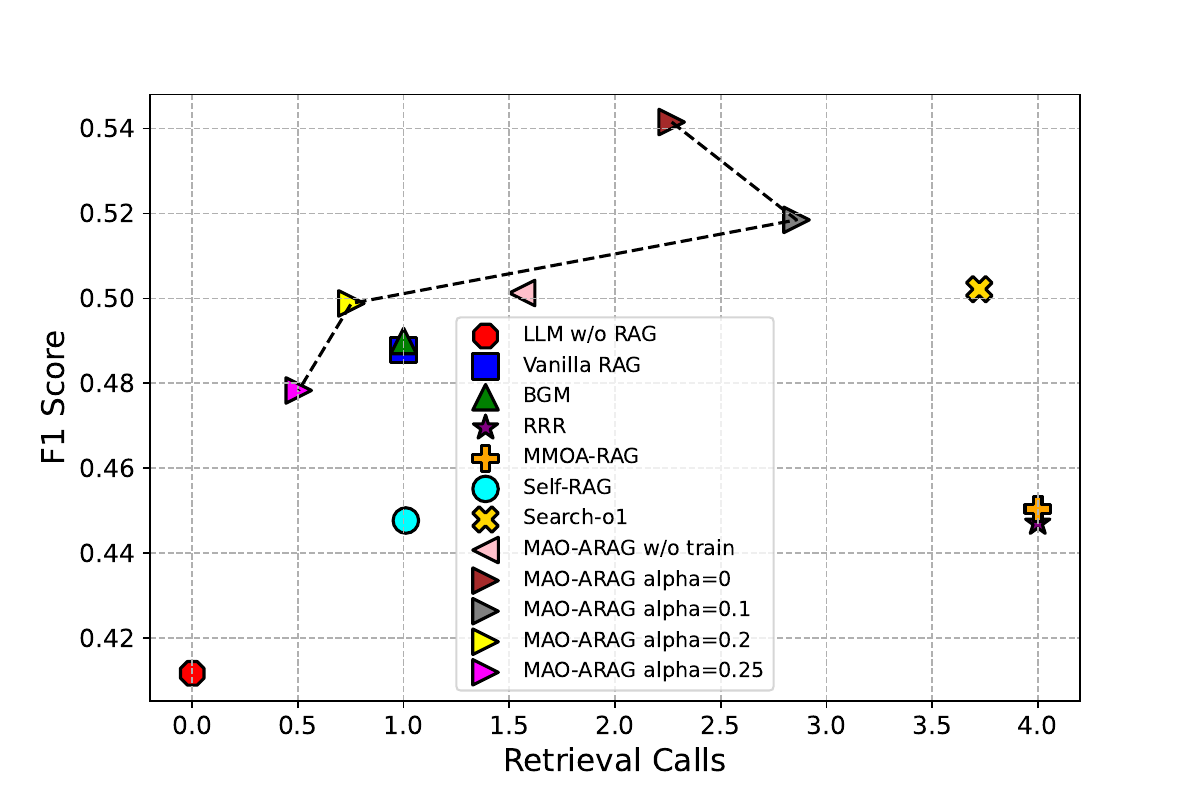}
  \caption{F1 score vs. Retrieval Calls}
  \label{f1_apitimes}
\end{subfigure}
\hfill
\begin{subfigure}{0.33\textwidth}
  \centering
  \includegraphics[width=\linewidth]{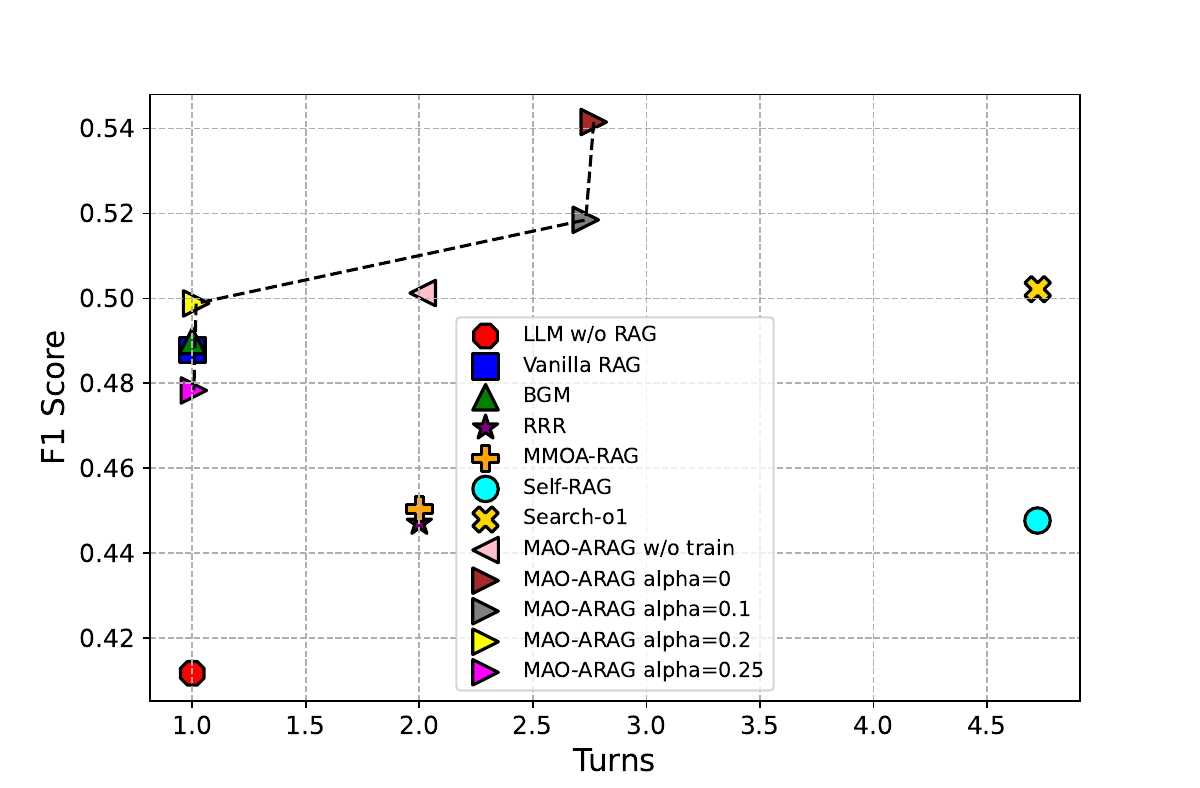}
  \caption{F1 score vs. Turns}
  \label{f1_turn_number}
\end{subfigure}

\caption{F1 score vs. cost metrics of different methods. All data in this figure is the average of NQ and HotpotQA datasets.}
\label{trade_off}
\end{figure*}

The F1 score is used to evaluate the quality of predicted answers $a_{\text{predicted}}$ across different methods, but the cost of generating these predictions is also important. In this study, we assessed three metrics related to prediction cost:
\begin{itemize}
    \item \textbf{Token Cost}: Represents the average cost of tokens consumed to answer predictions (in USD per query).
    \item \textbf{Retrieval Calls}: Indicates the average number of retrieval calls made (calls per query).
    \item \textbf{Turns}: The average number of turns required to complete a query (turns per query).
\end{itemize}

Figure \ref{trade_off} illustrates the relationship between the performance metric (F1 score) and the three prediction cost metrics.\footnote{Detail cost metrics can be seen in Appendix B.} A higher value on the horizontal axis signifies greater cost consumption, while a higher value on the vertical axis indicates better performance. Therefore, methods positioned closer to the top-left corner of the graphs achieve superior results with fewer resource expenditures. 

Our MAO-ARAG $\alpha=0$ achieved the highest F1 score, yet its cost metrics were not the highest. In Figure \ref{trade_off}, MAO-ARAG with different $\alpha$ forms a black dotted line, which is relatively close to the top-left corner, indicating that MAO-ARAG can achieve optimal performance at a relatively reasonable cost. Notably, although MAO-ARAG w/o train and Search-o1 have similar F1 scores, the cost metrics for MAO-ARAG w/o train are significantly lower than those for Search-o1. This suggests that our proposed architecture, which separates workflow planning and execution, inherently promotes more efficient resource use. While the costs for MAO-ARAG increase moderately after RL training compared to MAO-ARAG w/o train, its performance sees a substantial improvement of approximately 4\%. 

The performance of Search-r1 is somewhat inferior compared to our method. This discrepancy arises from the fact that in Search-r1, the implicit workflow planning and execution are tightly coupled, with all processes executed by a trainable agent based on an open-source LLM. The necessity for the model to be trainable, combined with the integration of planning and execution, results in suboptimal performance for Search-r1. This also highlights the advantages of our framework, which distinctly separates planning and execution while enabling the planner agent to be trainable.

\subsection{4.4\hspace{0.5em}Effect of Different Cost Weight $\alpha$ (RQ.3)}

\begin{figure}[t]
\centering
\begin{subfigure}{0.23\textwidth}
  \centering
  \includegraphics[width=\linewidth]{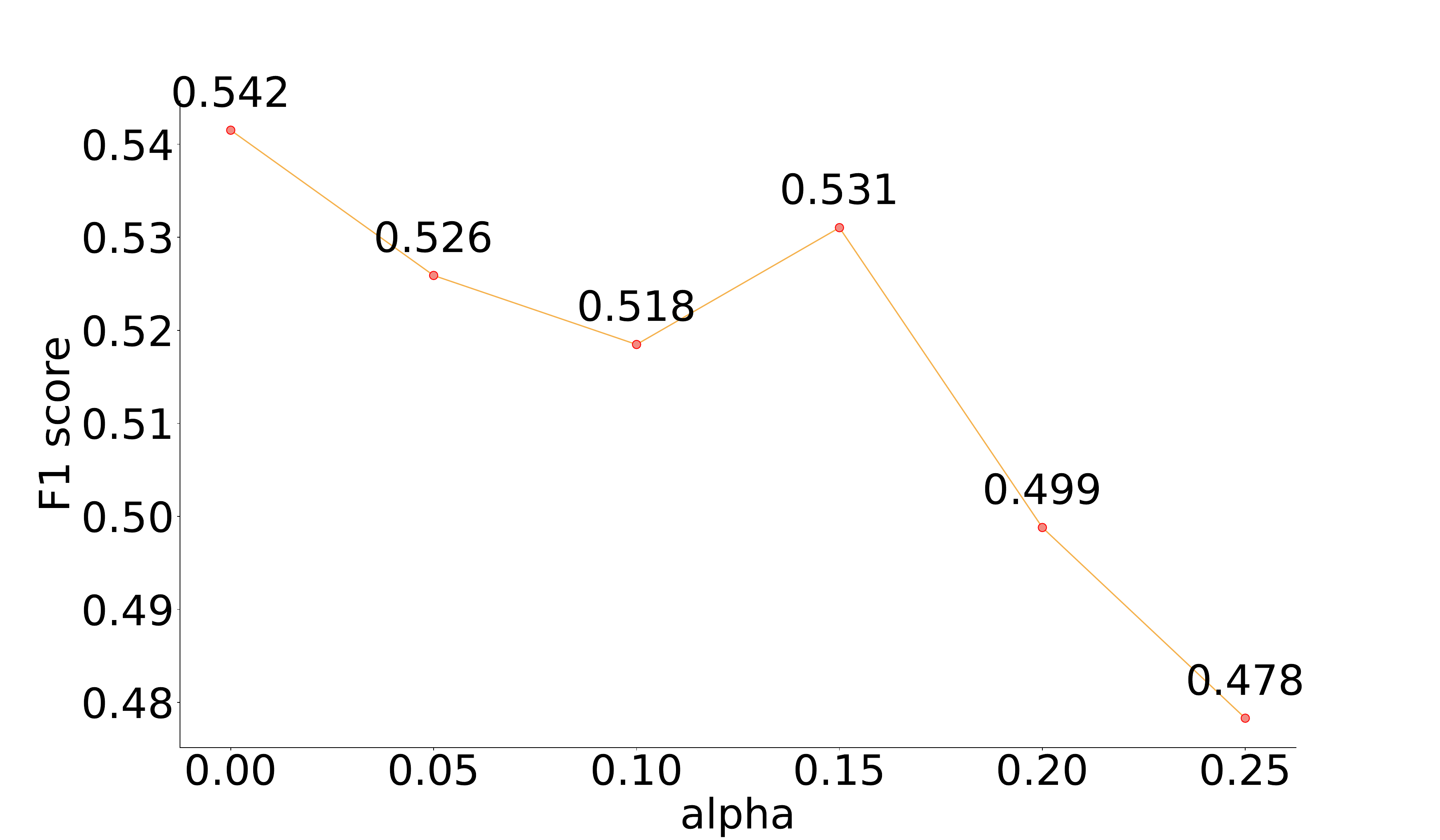}
  \caption{F1 Score vs. $\alpha$}
  \label{zhexian_f1}
\end{subfigure}
\hfill
\begin{subfigure}{0.23\textwidth}
  \centering
  \includegraphics[width=\linewidth]{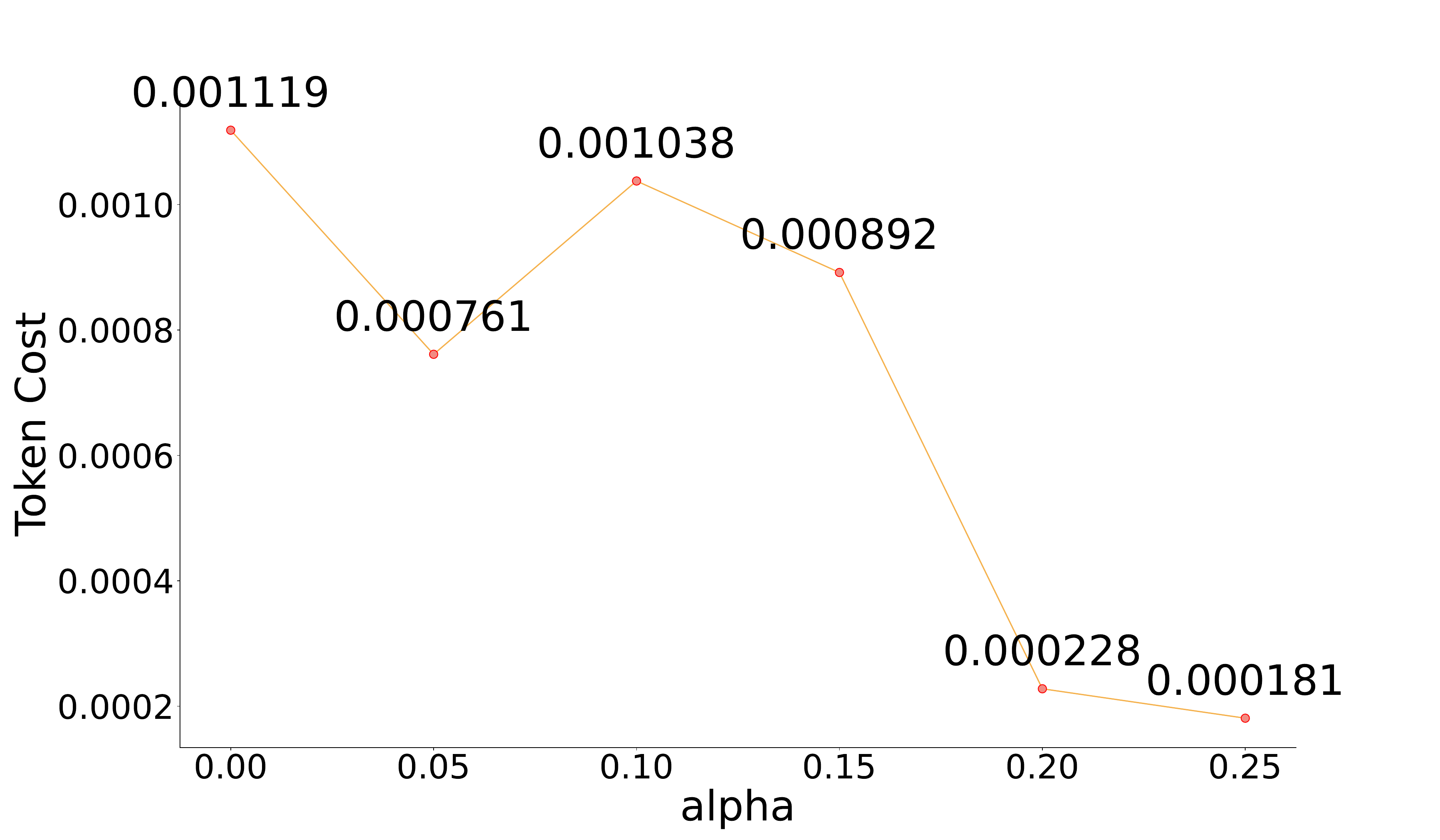}
  \caption{Token Cost vs. $\alpha$}
  \label{zhexian_cost}
\end{subfigure}
\hfill
\begin{subfigure}{0.23\textwidth}
  \centering
  \includegraphics[width=\linewidth]{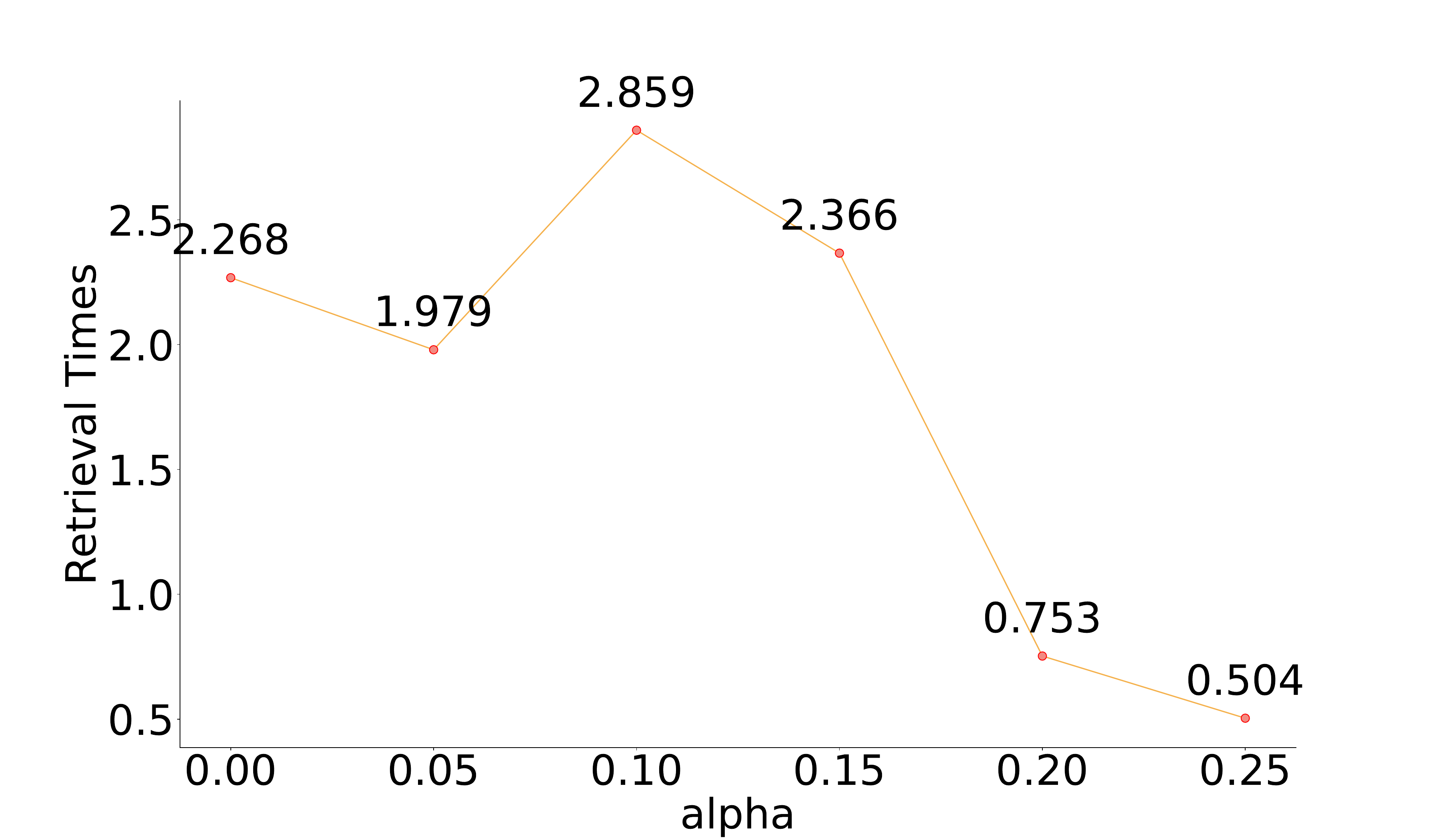}
  \caption{Retrieval Calls vs. $\alpha$}
  \label{zhexian_apitimes}
\end{subfigure}
\hfill
\begin{subfigure}{0.23\textwidth}
  \centering
  \includegraphics[width=\linewidth]{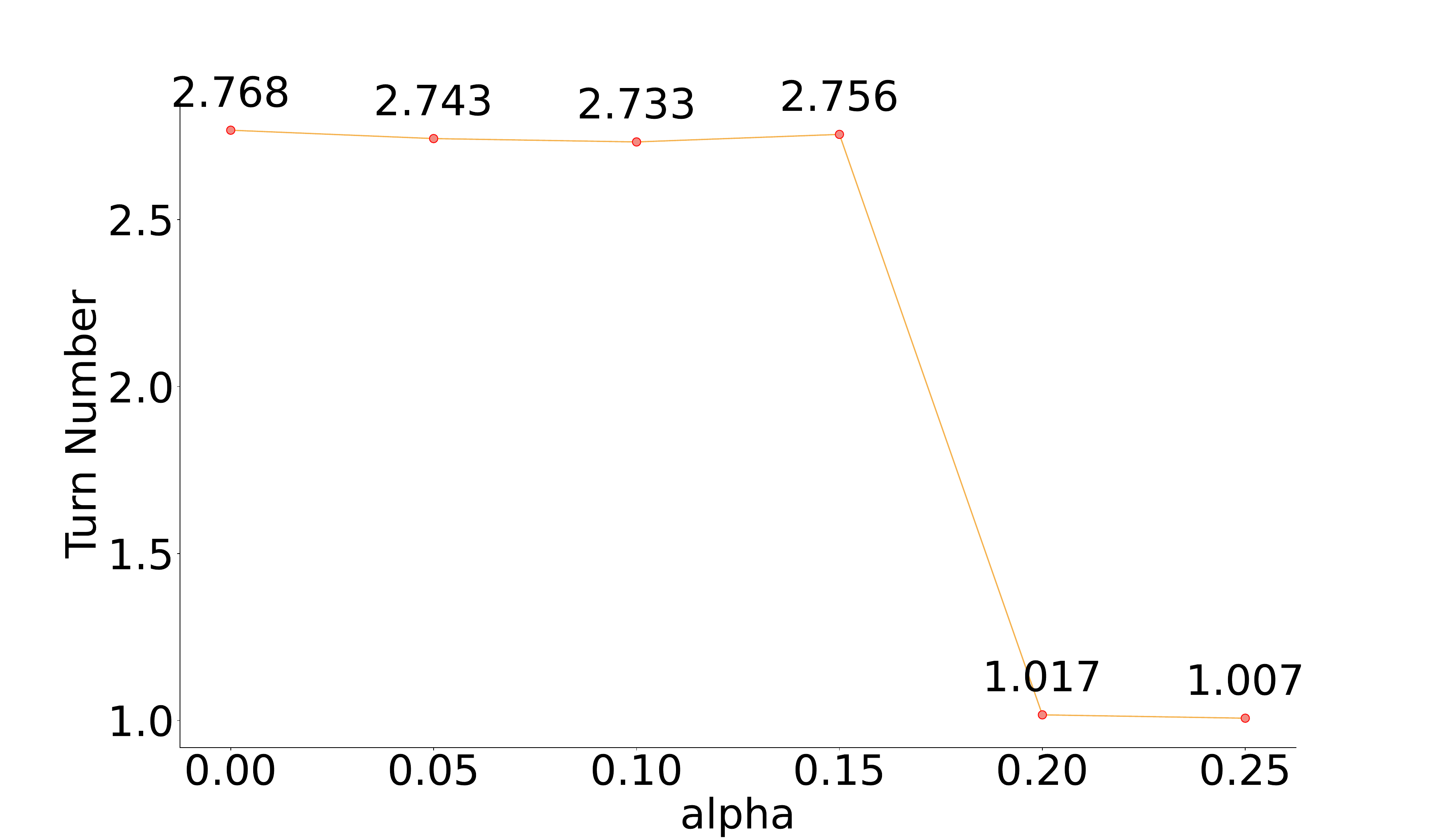}
  \caption{Turns vs. $\alpha$}
  \label{zhexian_turns}
\end{subfigure}

\caption{The metrics vs. $\alpha$ of different methods. All data in this figure is the average of NQ and HopotQA datasets.}
\label{different_alpha}
\end{figure}

In reinforcement learning, the ultimate strategy adopted by an agent is highly correlated with the reward function. Within our MAO-ARAG framework, the reward function is defined in Equation (\ref{reward fuction}), where the hyperparameter $\alpha$ governs the cost-based penalty term. By tuning the value of $\alpha$, we can achieve a balance between the effectiveness and the cost of the RAG pipeline. Theoretically, as $\alpha$ increases, the penalty on cost intensifies, which may degrade the planner's workflow performance while reducing the cost associated with obtaining answers (such as token cost, retriever call times, and latency). Conversely, a decrease in $\alpha$ enhances performance but incurs higher costs.

Figure \ref{different_alpha} illustrates the line graphs depicting the performance metric F1 score alongside three cost indicators under varying $\alpha$ values. It is evident that as $\alpha$ increases, there is a general decline in overall performance (F1 score), accompanied by a reduction in the three cost metrics due to the heightened penalty. Interestingly, when $\alpha$ exceeds 0.2, a rapid decline in various performance metrics occur. This phenomenon can be attributed to the fact that, as $\alpha$ increases, the cost-based penalty term in the reward function becomes disproportionately large. Consequently, the trained planner tends to generate overly simplistic workflows.

Additionally, it can be observed that the curves in Figure \ref{different_alpha} exhibit fluctuations. This may be due to the following reasons: (1) The limited number of executors defined may mean that the optimal workflow is composed of fewer executors, introducing significant uncertainty and causing fluctuations. (2) To simplify the definition of the Cost Penalty term $R_{CP}$ in Equation (\ref{CP}), we scaled the token cost $Token_{cost}$, turn numbers cost $Turn_{cost}$, and search engine call cost $\mathbb{I}(S)$ to a range of $[0, 1]$. However, there might be inherent weights among different cost penalty terms that ought to be considered. This coarse definition of the Cost Penalty term $R_{CP}$ could also contribute to the observed fluctuations, indicating a potential area for future refinement.

\subsection{4.5\hspace{0.5em}Smaller Planner and Different Executor Backbone (RQ.4)}

\begin{table}[t] \scriptsize
\caption{Smaller planner and other different backbones of executors. (Average metrics of NQ and HotpotQA datasets)}
\centering
\renewcommand{\arraystretch}{0.8}
\begin{tabular}{>{\centering\arraybackslash}m{2.0cm}
                >{\centering\arraybackslash}m{1.05cm}
                >{\centering\arraybackslash}m{1.05cm}
                >{\centering\arraybackslash}m{1.05cm}
                >{\centering\arraybackslash}m{1.15cm}}
\toprule
\textbf{Model \& Backbone} & \textbf{F1 Score} & \textbf{Token Cost} & \textbf{Retrieve Times} & \textbf{Turn Number} \\
\midrule
% \multicolumn{5}{c}{MAO-ARAG} \\
7B w/o train & 50.13 & 0.00064 & 1.56 & 2.02 \\
7B (+PPO) & 54.15 & 0.00112 & 2.27 & 2.77 \\
% \midrule
% 3B (+SFT) & 52.92 & 0.00102 & 2.13 & 2.60 \\
% 3B (+PPO) & 53.34 & 0.00108 & 2.22 & 2.72 \\
\midrule
\multicolumn{5}{c}{Smaller Planner} \\
1.5B (+SFT) & 53.81 & 0.00102 & 2.13 & 2.59 \\
1.5B (+PPO) & 53.91 & 0.00095 & 2.06 & 2.48 \\
% \midrule
0.5B (+SFT) & 53.64 & 0.00101 & 2.10 & 2.54 \\
0.5B (+PPO) & 53.92 & 0.00111 & 2.26 & 2.76 \\
\midrule
\multicolumn{5}{c}{Different Executor Backbone} \\
GPT-3.5-turbo & 48.08 & 0.00328 & 1.63 & 2.24 \\
GPT-4.1-nano & 47.43 & 0.00051 & 1.68 & 2.13 \\
\bottomrule
\end{tabular}
\label{different_model_api}
\end{table}

In this section, we explore the feasibility of using smaller models as planners and conduct preliminary experiments using alternative APIs as executor backbones.

We initiate our study by distilling the trained 7B planner model into 1.5B and 0.5B models using supervised fine-tuning (+SFT), followed by PPO training (+PPO). As shown in Table 
\ref{different_model_api}, the performance and cost metrics of these smaller planners (+SFT and +PPO) closely match those of the 7B model (+PPO). This indicates that larger planners can be effectively distilled into smaller models capable of performing the planner’s role just as well.

Furthermore, while previous experiments predominantly used GPT-4o-mini for the executor agents, we replaced this backbone with GPT-3.5-turbo and GPT-4.1-nano, respectively. From Table \ref{different_model_api}, we can see that both alternatives achieve similar F1 scores, albeit lower than the 7B (+PPO) using GPT-4o-mini as the executors’ backbone. However, due to the lower cost of the GPT-4.1-nano API, its token cost is only 0.00051, less than half of the 0.00112 incurred by the 7B (+PPO). Conversely, GPT-3.5-turbo, being an outdated model, not only results in a lower F1 score but also incurs a higher token cost.

The experiments in this section demonstrate that we can use smaller planners and more cost-effective APIs as executors’ backbones, achieving a more favorable balance between effectiveness and cost.

\section{5\hspace{1em}Conclusion}

In this paper, we proposed MAO-ARAG, a novel multi-agent orchestration framework for adaptive RAG in QA systems. MAO-ARAG dynamically constructs appropriate workflows for diverse queries, leveraging multiple executor agents, including query reformulation, document selection, and answer generation modules. These agents are orchestrated by a planner agent optimized using RL with a reward function that balances answer quality and cost metrics. Through extensive experiments on a variety of single-hop and multi-hop QA datasets, we demonstrated that MAO-ARAG outperforms existing RAG pipelines, achieving a better balance between effectiveness and cost.

Future work may focus on refining cost penalties to better balance performance and cost. We also plan to use multiple optional APIs simultaneously as executor backbones, aiming for better results at a lower cost.

\bibliography{aaai2026}

\clearpage

\section{Appendix}

\subsection{A\hspace{0.5em}How and why should the cost-based penalty terms be scaled?}
\label{How and why should the cost-based penalty terms be scaled?}

We define the cost-based penalty term $R_{CP}$ in the following Equation (Equal to Equation (\ref{CP})).

\begin{equation}\small
R_{CP} = Token_{\text{cost}} + Turn_{\text{cost}} + \mathbb{I}(S)
\end{equation}

We can see that $R_{CP}$ is composed of three parts: $Token_{\text{cost}}$, $Turn_{\text{cost}}$, and $\mathbb{I}(S)$. The range of values for these three parts varies significantly. In order to effectively optimize each component, we should normalize their values to approximately between 0 and 1.

Of course, in real-world scenarios, these three components can all be converted into actual money spent, and each component will have some proportional relationship in terms of cost. However, for the purpose of this study, we consider these three components to be equally important and optimize all three.

Next, we will introduce how to scale every term in this Equation.

\begin{itemize}
    \item $Token_{\text{cost}}$: Table \ref{Agent token cost (dollar per query)} presents the average token cost for each executor agent. Among the executable workflows, $QR, DS, AG, AS$ stands out as the most costly, with a token cost of approximately 6.02e-4 dollars per query. By scaling this value to 1.0, we can linearly adjust the token costs of all workflow types, denoted as $Token_{\text{cost}}$, to a range between 0 and 1.0.

    \begin{table}[h]\small
        \centering
        \caption{Agent token cost (per query)}
        \begin{tabular}{l c}
            \toprule
            \textbf{Executor Name} & \textbf{Token Cost} \\
            \midrule
            Query Decomposition Serial (QDS) & 0.91e-4 \\
            Query Decomposition Parallel (QDP) & 1.00e-4 \\
            Query Rewriter (QR) & 0.88e-4 \\
            Document Selector (DS) & 2.08e-4 \\
            Answer Generator (AG) & 1.58e-4 \\
            Answer Summarization (AS) & 1.48e-4 \\
            \bottomrule
        \end{tabular}
        \label{Agent token cost (dollar per query)}
    \end{table}

    \item $Turn_{\text{cost}}$: This penalty term is related to latency. The more rounds a query requires, the longer it will take to complete, resulting in higher latency. 
    
    Among all the executors, only QDS and QDP incur additional subsequent rounds. Since QDP decomposes the original question into multiple sub-questions that can be searched in parallel, it results in one additional round, as all sub-questions can be processed simultaneously. On the other hand, QDS decomposes the original question into sub-questions that must be searched sequentially, meaning that each subsequent sub-question requires the answer from the previous one to proceed to the next search step. Therefore, QDS results in additional rounds equal to the number of sub-questions. As we limit the maximum number of sub-questions to four rounds, we have normalized the turncost for all executors to a range between 0 and 1, as shown in the table below:

    \begin{table}[h]\small
        \centering
        \caption{$Turn_{\text{cost}}$ for different workflows:}
        \begin{tabular}{l c}
            \toprule
            \textbf{Workflow} & \textbf{Turn Cost} \\
            \midrule
            QDS & 0.25 \\
            QDP & 0.25, 0.5, 0.75, 1.0 \\
            QR, RA, AG & 0 \\
            RA, DS, AG & 0 \\
            AS & 0 \\
            Other Workflows & 0 \\
            \bottomrule
        \end{tabular}
        \label{turn cost for different workflows}
    \end{table}

    \item $\mathbb{I}(S)$: This is a penalty term related to the number of retrieval model calls. Since the cost of calling the search engine’s API is relatively high, this penalty term is included to encourage the planner to minimize the expenses associated with using the search engine. Specifically, if the workflow output by the planner includes a Retrieval Agent (RA), we have $\mathbb{I}(S)=1$; otherwise, we set $\mathbb{I}(S)=0$.

\end{itemize}

\subsection{B\hspace{0.5em}Detailed Values of Different Cost Metrics}
\label{Detailed Values of Different Cost Metrics}

In Tables \ref{tab:token_costs}, \ref{tab:api_times}, and \ref{tab:turn_num}, we present the detailed values of three cost-based metrics—token cost, retrieval call times, and turn number—across various methods on different datasets (averaged per query). Tables \ref{tab:token_costs}, \ref{tab:api_times}, and \ref{tab:turn_num} correspond to the three subplots in Figure \ref{trade_off}. It is important to note that Figure \ref{trade_off} shows the averages for the NQ and HotpotQA datasets, whereas Tables \ref{tab:token_costs}, \ref{tab:api_times}, and \ref{tab:turn_num} present data for all datasets.

\subsection{C\hspace{0.5em}Prompt Details}

Figures~\ref{QDS_prompt} through~\ref{AS_prompt} present the detailed prompt templates used by each agent in the MAO-ARAG framework. Specifically, Figure~\ref{QDS_prompt} shows the prompt for the \textbf{Query Decomposition Serial (QDS)} agent. Figure~\ref{QDP_prompt} presents the prompt used by the \textbf{Query Decomposition Parallel (QDP)} agent. Figure~\ref{QR_prompt} provides the prompt for the \textbf{Query Rewriter (QR)} agent. Figure~\ref{DS_prompt} illustrates the prompt for the \textbf{Document Selector (DS)} agent. Figure~\ref{AG_prompt} shows the prompt for the \textbf{Answer Generator (AG)} agent. Finally, Figure~\ref{AS_prompt} displays the prompt for the \textbf{Answer Summarization (AS)} agent.

\subsection{D\hspace{0.5em}Limitations}
\label{Limitations}

Since our MAO-ARAG requires training the planner agent using RL, the planner’s backbone model must effectively follow instructions and have a decent initial ability to plan workflows. In our experiments, it was not possible to train directly based on Qwen2.5-0.5B-Instruct and Qwen2.5-1.5B-Instruct because models of this size have issues with instruction-following capabilities.

% \clearpage

\subsection{E\hspace{0.5em}Case Study}

To further demonstrate the effectiveness and adaptability of the proposed \textsc{MAO-ARAG} framework, we present a case study illustrating how our system dynamically generates tailored workflows for different types of queries. Each case is structured as follows: we begin with a user query and its corresponding golden answer. Then, we detail each interaction turn within the MAO-ARAG framework. At each turn, the \textbf{planner} selects a workflow by orchestrating a set of \textbf{executor}. The selected workflow is then executed, and the resulting context—including sub-questions, intermediate answers, and retrieved documents—is accumulated. The process continues iteratively until a final answer is produced, at which point it is compared against the golden answer.

We present four representative cases to highlight MAO-ARAG’s ability to adaptively choose between simple and complex workflows based on query demands.

\paragraph{Case 1: Single-Turn Answer Generation}

\textbf{Query: Is aluminium a ferrous or non ferrous metal?} (From NQ)

This is a straightforward factual question that can be confidently answered from the language model's internal knowledge. The planner correctly identifies that no retrieval or decomposition is necessary and directly selects the \textbf{AG} module in a single turn. The model produces the correct answer \textbf{non-ferrous}, which matches the golden answer. This case exemplifies MAO-ARAG's ability to avoid unnecessary computations and costs for simple queries.

% Create a tcolorbox for the case study
\begin{tcolorbox}[breakable, title=Single-Turn Answer Generation]

\begin{itemize} \scriptsize

    \item \textbf{Initial question $q$}:
    \begin{itemize}
        \item Is aluminium a ferrous or non ferrous metal?
    \end{itemize}
    % \noindent\rule{\dimexpr\linewidth-4mm\relax}{0.4pt} % Horizontal line

    \item \textbf{Golden answer} $Ans_{\text{golden}}$:
    \begin{itemize}
        \item non-ferrous
    \end{itemize}

    \item \textbf{Turn 0}:
    \begin{itemize}
        \item \textbf{Planner: AG}

        \item \textbf{Executor: AG}
        \item \textbf{Context:}
            \begin{itemize}
                \item \textbf{Question:} Is aluminium a ferrous or non ferrous metal?
                \item \textbf{Answer:} non ferrous metal
            \end{itemize}
    \end{itemize}

    \item \textbf{Predicted answer} $Ans_{\text{predict}}$:
    \begin{itemize}
        \item non ferrous metal {\color{green}\cmark}
    \end{itemize}

    \end{itemize}

\end{tcolorbox}

\paragraph{Case 2: Single-Turn Retrieval-Augmented Generation}

\textbf{Query: Who was the editor of the journal Jugantor published in the time of Swadeshi movement?} (From NQ)

This question requires external knowledge not reliably stored in the model’s parameters. The planner selects a one-turn plan involving retrieval followed by generation (\textbf{RA} $\rightarrow$ \textbf{AG}). The retrieved documents contain relevant historical context, enabling the model to correctly identify \textbf{Bhupendranath Dutt} as the editor. Notably, if the model attempted to answer without retrieval, it produced incorrect or hallucinated content. This demonstrates MAO-ARAG’s capability to recognize knowledge gaps and invoke retrieval when necessary.

\begin{tcolorbox}[breakable, title=Single-Turn Retrieval-Augmented Generation]

\begin{itemize} \scriptsize

    \item \textbf{Initial question $q$}:
    \begin{itemize}
        \item Who was the editor of the journal jugantor published in the time of swadeshi movement?
    \end{itemize}

    \item \textbf{Golden answer} $Ans_{\text{golden}}$:
    \begin{itemize}
        \item Bhupendranath Dutt
    \end{itemize}

    \item \textbf{Turn 0}:
    \begin{itemize}
        \item \textbf{Planner: RA, AG}

        \item \textbf{Executor: RA}
        \item \textbf{Context:}
            \begin{itemize}
                \item \textbf{Question:} Who was the editor of the journal jugantor published in the time of swadeshi movement?
                \item \textbf{Documents:}
                    \begin{itemize}
                        \item Jugantar Patrika () was a Bengali revolutionary newspaper founded in 1906 in Calcutta by Barindra Kumar Ghosh, Abhinash Bhattacharya and \textbf{Bhupendranath Dutt}. A political weekly, it was founded in March 1906 and served as the propaganda organ for the nascent revolutionary organisation "Anushilan Samiti"...
                    \end{itemize}
        \end{itemize}
        \item \textbf{Executor: AG}
        \item \textbf{Context:}
            \begin{itemize}
                \item \textbf{Question:} Who was the editor of the journal jugantor published in the time of swadeshi movement?
                \item \textbf{Documents:}
                    \begin{itemize}
                        \item Jugantar Patrika () was a Bengali revolutionary newspaper founded in 1906 in Calcutta by Barindra Kumar Ghosh, Abhinash Bhattacharya and \textbf{Bhupendranath Dutt}. A political weekly, it was founded in March 1906 and served as the propaganda organ for the nascent revolutionary organisation "Anushilan Samiti"...
                    \end{itemize}
                \item \textbf{Answer:} Bhupendranath Dutt
            \end{itemize}

    \end{itemize}

     \item \textbf{Predicted answer} Without RAG:
    \begin{itemize}
        \item Bepin Chandra Pal {\textcolor{red}{\ding{55}}}
    \end{itemize}

    \item \textbf{Predicted answer} With RAG (RA + AG) $Ans_{\text{predict}}$:
    \begin{itemize}
        \item Bhupendranath Dutt {\textcolor{green}{\ding{51}}}
    \end{itemize}

\end{itemize}

\end{tcolorbox}

\paragraph{Case 3: Multi-Turn Parallel Query Decomposition}

\textbf{Query: Which performance act has a higher instrument to person ratio, Badly Drawn Boy or Wolf Alice?} (From HotpotQA)

This is a complex comparative question requiring reasoning over multiple independent facts. The planner decomposes the main query into four focused sub-questions: (1) How many members are in the performance act Badly Drawn Boy?, (2) How many instruments are typically used in a performance by Badly Drawn Boy?, (3) How many members are in the performance act Wolf Alice?, and (4) How many instruments are typically used in a performance by Wolf Alice? These sub-queries are processed in parallel, with the planner independently determining whether external retrieval is necessary for each. Once all intermediate results are obtained, the planner invokes the \textbf{AS} module to compute the instrument-to-person ratios and synthesize a comparative answer. This case illustrates MAO-ARAG’s ability to perform fine-grained parallel reasoning and adapt retrieval strategies to the needs of each sub-question.

\begin{tcolorbox}[breakable, title=Multi-Turn Parallel Query Decomposition]

\begin{itemize} \scriptsize

    \item \textbf{Initial question $q$}:
    \begin{itemize}
        \item Which performance act has a higher instrument to person ratio, Badly Drawn Boy or Wolf Alice?
    \end{itemize}

    \item \textbf{Golden answer} $Ans_{\text{golden}}$:
    \begin{itemize}
        \item Badly Drawn Boy
    \end{itemize}

    \item \textbf{Turn 0}:
    \begin{itemize}
        \item \textbf{Planner: QDP}

        \item \textbf{Executor: QDP}
        \item \textbf{Context:}
            \begin{itemize}
                \item \textbf{Question:} Which performance act has a higher instrument to person ratio, Badly Drawn Boy or Wolf Alice?
                \item \textbf{Sub-questions $subq$}:
                \begin{itemize}
                    \item Sub-question 1: How many members are in the performance act Badly Drawn Boy?
                    \item Sub-question 2: How many instruments are typically used in a performance by Badly Drawn Boy?
                    \item Sub-question 3: How many members are in the performance act Wolf Alice?
                    \item Sub-question 4: How many instruments are typically used in a performance by Wolf Alice?
                \end{itemize}
            \end{itemize}
    \end{itemize}
    
    \item \textbf{Turn 1}:
    \begin{itemize}
        \item \textbf{planner: AG (parallel for sub-question 1)}
        \item \textbf{Executor: AG}
        \item \textbf{Context:}
            \begin{itemize}
                \item \textbf{Question:} Which performance act has a higher instrument to person ratio, Badly Drawn Boy or Wolf Alice?
                \item \textbf{Sub-questions $subq$:}
                \begin{itemize}
                    \item Sub-question 1: How many members are in the performance act Badly Drawn Boy?
                    \item Sub-question 2: How many instruments are typically used in a performance by Badly Drawn Boy?
                    \item Sub-question 3: How many members are in the performance act Wolf Alice?
                    \item Sub-question 4: How many instruments are typically used in a performance by Wolf Alice?
                \end{itemize}
                \item \textbf{Documents:}
                \begin{itemize}
                        \item Documents for Sub-question 1: None
                \end{itemize}
                \item \textbf{Sub-answers $suba$:}
                    \begin{itemize}
                        \item Sub-answer 1: One member
                    \end{itemize}
            \end{itemize}

        \item \textbf{planner: AG (parallel for sub-question 2)}
        \item \textbf{Executor: AG}
        
        \item \textbf{Context:}
            \begin{itemize}
                \item \textbf{Question:} Which performance act has a higher instrument to person ratio, Badly Drawn Boy or Wolf Alice?
                \item \textbf{Sub-questions $subq$:}
                \begin{itemize}
                    \item Sub-question 1: How many members are in the performance act Badly Drawn Boy?
                    \item Sub-question 2: How many instruments are typically used in a performance by Badly Drawn Boy?
                    \item Sub-question 3: How many members are in the performance act Wolf Alice?
                    \item Sub-question 4: How many instruments are typically used in a performance by Wolf Alice?
                \end{itemize}
                \item \textbf{Documents:}
                \begin{itemize}
                        \item Documents for Sub-question 1: None
                        \item Documents for Sub-question 2: None
                \end{itemize}
                \item \textbf{Sub-answers $suba$:}
                    \begin{itemize}
                        \item Sub-answer 1: One member
                        \item Sub-answer 2: Typically four instruments are used in a performance by Badly Drawn Boy.
                    \end{itemize}
                
            \end{itemize}

        \item \textbf{planner: AG (parallel for sub-question 3)}
        \item \textbf{Executor: AG}

        \item \textbf{Context:}
            \begin{itemize}
                \item \textbf{Question:} Which performance act has a higher instrument to person ratio, Badly Drawn Boy or Wolf Alice?
                \item \textbf{Sub-questions $subq$:}
                \begin{itemize}
                    \item Sub-question 1: How many members are in the performance act Badly Drawn Boy?
                    \item Sub-question 2: How many instruments are typically used in a performance by Badly Drawn Boy?
                    \item Sub-question 3: How many members are in the performance act Wolf Alice?
                    \item Sub-question 4: How many instruments are typically used in a performance by Wolf Alice?
                \end{itemize}
                \item \textbf{Documents:}
                \begin{itemize}
                        \item Documents for Sub-question 1: None
                        \item Documents for Sub-question 2: None
                        \item Documents for Sub-question 3: None
                \end{itemize}
                
                \item \textbf{Sub-answers $suba$:}
                    \begin{itemize}
                        \item Sub-answer 1: One member
                        \item Sub-answer 2: Typically four instruments are used in a performance by Badly Drawn Boy.
                        \item Sub-answer 3: Four
                    \end{itemize}
                
            \end{itemize}

        \item \textbf{planner: RA, AG (parallel for sub-question 4)}

        \item \textbf{Executor: RA}

        \item \textbf{Context:}
        \begin{itemize}
            \item \textbf{Question:} Which performance act has a higher instrument to person ratio, Badly Drawn Boy or Wolf Alice?
            \item \textbf{Sub-questions $subq$:}
            \begin{itemize}
                \item Sub-question 1: How many members are in the performance act Badly Drawn Boy?
                \item Sub-question 2: How many instruments are typically used in a performance by Badly Drawn Boy?
                \item Sub-question 3: How many members are in the performance act Wolf Alice?
                \item Sub-question 4: How many instruments are typically used in a performance by Wolf Alice?
            \end{itemize}
            \item \textbf{Documents:}
                \begin{itemize}
                        \item Documents for Sub-question 1: None
                        \item Documents for Sub-question 2: None
                        \item Documents for Sub-question 3: None
                        \item Documents for Sub-question 4: construct her musical universe. She created her debut album Coppé entirely on a TEAC Reel-to-Reel, transitioning through Yamaha’s DX7 to the Nord Lead and \textbf{her current instrument of choice, Teenage Engineering’s versatile OP-1. Outside the synth world Coppé employs unconventional instruments including the 5-octave mbira, nail violin and meatgrinder.} Coppé characteristically processes her vocals to create a broad range of effects that she likens to “angels whispering,” “colors of the wind,” and “orange sorbet sunsets” with equipment including the Digitech 300 and TC helicon. Coppé co-creates with some of the electronic music/IDM scene's most influential artists. These include Plaid, Kettel,...
                \end{itemize}
                
            \item \textbf{Sub-answers $suba$:}
                \begin{itemize}
                    \item Sub-answer 1: One member
                    \item Sub-answer 2: Typically four instruments are used in a performance by Badly Drawn Boy.
                    \item Sub-answer 3: Four
                \end{itemize}

        \end{itemize}
    
        \item \textbf{Executor: AG}

        \item \textbf{Context:}
        \begin{itemize}
            \item \textbf{Question:} Which performance act has a higher instrument to person ratio, Badly Drawn Boy or Wolf Alice?
            \item \textbf{Sub-questions $subq$:}
            \begin{itemize}
                \item Sub-question 1: How many members are in the performance act Badly Drawn Boy?
                \item Sub-question 2: How many instruments are typically used in a performance by Badly Drawn Boy?
                \item Sub-question 3: How many members are in the performance act Wolf Alice?
                \item Sub-question 4: How many instruments are typically used in a performance by Wolf Alice?
            \end{itemize}
            \item \textbf{Documents:}
                \begin{itemize}
                        \item Documents for Sub-question 1: None
                        \item Documents for Sub-question 2: None
                        \item Documents for Sub-question 3: None
                        \item Documents for Sub-question 4: construct her musical universe. She created her debut album Coppé entirely on a TEAC Reel-to-Reel, transitioning through Yamaha’s DX7 to the Nord Lead and \textbf{her current instrument of choice, Teenage Engineering’s versatile OP-1. Outside the synth world Coppé employs unconventional instruments including the 5-octave mbira, nail violin and meatgrinder.} Coppé characteristically processes her vocals to create a broad range of effects that she likens to “angels whispering,” “colors of the wind,” and “orange sorbet sunsets” with equipment including the Digitech 300 and TC helicon. Coppé co-creates with some of the electronic music/IDM scene's most influential artists. These include Plaid, Kettel,...
                \end{itemize}
            \item \textbf{Sub-answers $suba$:}
                \begin{itemize}
                    \item Sub-answer 1: One member
                    \item Sub-answer 2: Typically four instruments are used in a performance by Badly Drawn Boy.
                    \item Sub-answer 3: Four
                    \item Sub-answer 4: Four instruments
                \end{itemize}
        \end{itemize}
    \end{itemize}

    \item \textbf{Turn 2}:
    \begin{itemize}
        \item \textbf{planner: AS}
        \item \textbf{Executor: AS}
        \item \textbf{Context:}
        \begin{itemize}
            \item \textbf{Question:} Which performance act has a higher instrument to person ratio, Badly Drawn Boy or Wolf Alice?
            \item \textbf{Sub-questions $subq$:}
            \begin{itemize}
                \item Sub-question 1: How many members are in the performance act Badly Drawn Boy?
                \item Sub-question 2: How many instruments are typically used in a performance by Badly Drawn Boy?
                \item Sub-question 3: How many members are in the performance act Wolf Alice?
                \item Sub-question 4: How many instruments are typically used in a performance by Wolf Alice?
            \end{itemize}
            \item \textbf{Documents:}
                \begin{itemize}
                        \item Documents for Sub-question 1: None
                        \item Documents for Sub-question 2: None
                        \item Documents for Sub-question 3: None
                        \item Documents for Sub-question 4: construct her musical universe. She created her debut album Coppé entirely on a TEAC Reel-to-Reel, transitioning through Yamaha’s DX7 to the Nord Lead and \textbf{her current instrument of choice, Teenage Engineering’s versatile OP-1. Outside the synth world Coppé employs unconventional instruments including the 5-octave mbira, nail violin and meatgrinder.} Coppé characteristically processes her vocals to create a broad range of effects that she likens to “angels whispering,” “colors of the wind,” and “orange sorbet sunsets” with equipment including the Digitech 300 and TC helicon. Coppé co-creates with some of the electronic music/IDM scene's most influential artists. These include Plaid, Kettel,...
                \end{itemize}
            \item \textbf{Sub-answers $suba$:}
                \begin{itemize}
                    \item Sub-answer 1: One member
                    \item Sub-answer 2: Typically four instruments are used in a performance by Badly Drawn Boy.
                    \item Sub-answer 3: Four
                    \item Sub-answer 4: Four instruments
                \end{itemize}
            \item \textbf{Answer:} Badly Drawn Boy
        \end{itemize}
    \end{itemize}

    \item \textbf{Predicted final answer}  $Ans_{\text{predict}}$:
    \begin{itemize}
        \item Badly Drawn Boy {\textcolor{green}{\ding{51}}}
    \end{itemize}

\end{itemize}

\end{tcolorbox}

\paragraph{Case 4: Multi-Turn Sequential Query Decomposition} 

\textbf{Query:} \textbf{Ralph Hefferline was a psychology professor at a university that is located in what city?} (From HotpotQA)

This is a compositional question requiring sequential reasoning. The planner decomposes the query into two dependent sub-questions: (1) At which university was Ralph Hefferline a psychology professor?, followed by (2) In what city is this university located? The second sub-question is dynamically rewritten based on the first answer. After each planning phase, the appropriate modules are invoked (\textbf{RA} and \textbf{AG} as needed), and the answers are accumulated. Once both sub-questions are resolved, the \textbf{AS} module generates the final answer. This case showcases MAO-ARAG’s ability to handle sequential dependencies through multi-turn planning and sub-question reformulation.

\begin{tcolorbox}[breakable, title=Multi-Turn Sequential Query Decomposition]

\begin{itemize} \scriptsize

    \item \textbf{Initial question $q$}:
    \begin{itemize}
        \item Ralph Hefferline was a psychology professor at a university that is located in what city?
    \end{itemize}

    \item \textbf{Golden answer} $Ans_{\text{golden}}$:
    \begin{itemize}
        \item New York City
    \end{itemize}

    \item \textbf{Turn 0}:
    \begin{itemize}
        \item \textbf{Planner: QDS}

        \item \textbf{Executor: QDS}
        \item \textbf{Context:}
            \begin{itemize}
                \item \textbf{Question:} Ralph Hefferline was a psychology professor at a university that is located in what city?
                \item \textbf{Sub-questions $subq$}:
                \begin{itemize}
                    \item Sub-question 1: At which university was Ralph Hefferline a psychology professor?
                    \item Sub-question 2: In what city is this university located?
                \end{itemize}
            \end{itemize}
    \end{itemize}

    \item \textbf{Turn 1}:
    \begin{itemize}
        \item \textbf{Planner: RA, AG}

        \item \textbf{Executor: RA}
        \item \textbf{Context:}
            \begin{itemize}
                \item \textbf{Question:} Ralph Hefferline was a psychology professor at a university that is located in what city?
                \item \textbf{Sub-questions $subq$}:
                \begin{itemize}
                    \item Sub-question 1: At which university was Ralph Hefferline a psychology professor?
                    \item Sub-question 2: In what city is this university located?
                \end{itemize}

                \item \textbf{Documents:}
                \begin{itemize}
                        \item Documents for Sub-question 1: of psychology. Ralph Hefferline Ralph Franklin Hefferline (15 February 1910 in Muncie, Indiana – 16 March 1974) was a psychology professor at \textbf{Columbia University}. Hefferline became a patient of Fritz Perls around 1946. He joined a small training group led by Perls in 1948 in New York, and went on to contribute a chapter to the book which defined Gestalt Therapy, "Gestalt Therapy, Excitement and Growth in the Human Personality", co-authored by Perls, Paul Goodman and Hefferline, published in 1951. He was the third and junior author and provided the section containing practical exercises. He went on to join the...
                \end{itemize}
            \end{itemize}
            
        \item \textbf{Executor: AG}
        \item \textbf{Context:}
            \begin{itemize}
                \item \textbf{Question:} Ralph Hefferline was a psychology professor at a university that is located in what city?
                \item \textbf{Sub-questions $subq$}:
                \begin{itemize}
                    \item Sub-question 1: At which university was Ralph Hefferline a psychology professor?
                    \item Sub-question 2: In what city is this university located?
                \end{itemize}

                \item \textbf{Documents:}
                \begin{itemize}
                        \item Documents for Sub-question 1: of psychology. Ralph Hefferline Ralph Franklin Hefferline (15 February 1910 in Muncie, Indiana – 16 March 1974) was a psychology professor at \textbf{Columbia University}. Hefferline became a patient of Fritz Perls around 1946. He joined a small training group led by Perls in 1948 in New York, and went on to contribute a chapter to the book which defined Gestalt Therapy, "Gestalt Therapy, Excitement and Growth in the Human Personality", co-authored by Perls, Paul Goodman and Hefferline, published in 1951. He was the third and junior author and provided the section containing practical exercises. He went on to join the...
                \end{itemize}
                \item \textbf{Sub-answers $suba$:}
                \begin{itemize}
                    \item Sub-answer 1: Columbia University
                \end{itemize}
                
            \end{itemize}
        
    \end{itemize}

    \item \textbf{Turn 2}:
    \begin{itemize}
        \item \textbf{Planner: AG}
        \item \textbf{Executor: AG}
        \item \textbf{Context:}
            \begin{itemize}
                \item \textbf{Question:} Ralph Hefferline was a psychology professor at a university that is located in what city?
                \item \textbf{Sub-questions $subq$}:
                \begin{itemize}
                    \item Sub-question 1: At which university was Ralph Hefferline a psychology professor?
                    \item Sub-question 2: In what city is this university located?
                \end{itemize}

                \item \textbf{Documents:}
                \begin{itemize}
                        \item Documents for Sub-question 1: of psychology. Ralph Hefferline Ralph Franklin Hefferline (15 February 1910 in Muncie, Indiana – 16 March 1974) was a psychology professor at \textbf{Columbia University}. Hefferline became a patient of Fritz Perls around 1946. He joined a small training group led by Perls in 1948 in New York, and went on to contribute a chapter to the book which defined Gestalt Therapy, "Gestalt Therapy, Excitement and Growth in the Human Personality", co-authored by Perls, Paul Goodman and Hefferline, published in 1951. He was the third and junior author and provided the section containing practical exercises. He went on to join the...
                        \item Documents for Sub-question 2: None
                \end{itemize}
                \item \textbf{Sub-answers $suba$:}
                \begin{itemize}
                    \item Sub-answer 1: Columbia University
                    \item Sub-answer 1: New York City
                \end{itemize}
                
            \end{itemize}
        
    \end{itemize}
    
    \item \textbf{Turn 3}:
        \begin{itemize}
        \item \textbf{Planner: AS}
        \item \textbf{Executor: AS}
        \item \textbf{Context:}
            \begin{itemize}
                \item \textbf{Question:} Ralph Hefferline was a psychology professor at a university that is located in what city?
                \item \textbf{Sub-questions $subq$}:
                \begin{itemize}
                    \item Sub-question 1: At which university was Ralph Hefferline a psychology professor?
                    \item Sub-question 2: In what city is this university located?
                \end{itemize}

                \item \textbf{Documents:}
                \begin{itemize}
                        \item Documents for Sub-question 1: of psychology. Ralph Hefferline Ralph Franklin Hefferline (15 February 1910 in Muncie, Indiana – 16 March 1974) was a psychology professor at \textbf{Columbia University}. Hefferline became a patient of Fritz Perls around 1946. He joined a small training group led by Perls in 1948 in New York, and went on to contribute a chapter to the book which defined Gestalt Therapy, "Gestalt Therapy, Excitement and Growth in the Human Personality", co-authored by Perls, Paul Goodman and Hefferline, published in 1951. He was the third and junior author and provided the section containing practical exercises. He went on to join the...
                        \item Documents for Sub-question 2: None
                \end{itemize}
                \item \textbf{Sub-answers $suba$:}
                \begin{itemize}
                    \item Sub-answer 1: Columbia University
                    \item Sub-answer 1: New York City
                \end{itemize}
                \item \textbf{Answer:} New York City
            \end{itemize}
        
    \end{itemize}

    \item \textbf{Predicted final answer}  $Ans_{\text{predict}}$:
    \begin{itemize}
        \item New York City {\textcolor{green}{\ding{51}}}
    \end{itemize}

\end{itemize}

\end{tcolorbox}

% \subsection{Detailed Values of Different Cost Metrics}
% \label{Detailed Values of Different Cost Metrics}

% In Tables \ref{tab:token_costs}, \ref{tab:api_times}, and \ref{tab:turn_num}, we present the detailed values of three cost-based metrics—token cost, retrieval call times, and turn number—across various methods on different datasets (averaged per query). Tables \ref{tab:token_costs}, \ref{tab:api_times}, and \ref{tab:turn_num} correspond to the three subplots in Figure \ref{trade_off}. It is important to note that Figure \ref{trade_off} shows the averages for the NQ and HotpotQA datasets, whereas Tables \ref{tab:token_costs}, \ref{tab:api_times}, and \ref{tab:turn_num} present data for all datasets.

\begin{table*}[t] \small
\caption{Average token cost per query (in milli-USD, mUSD) of various methods across datasets.}
\centering
\begin{tabular}{>{\centering\arraybackslash}m{4.15cm}
                >{\centering\arraybackslash}m{1.2cm}
                >{\centering\arraybackslash}m{1.2cm}
                >{\centering\arraybackslash}m{1.2cm}
                >{\centering\arraybackslash}m{1.2cm}
                >{\centering\arraybackslash}m{1.2cm}
                >{\centering\arraybackslash}m{1.2cm}
                >{\centering\arraybackslash}m{1.2cm}
                >{\centering\arraybackslash}m{1.4cm}}
\toprule
\textbf{Methods} & \textbf{NQ} & \textbf{PopQA} & \textbf{AmbigQA} & \textbf{HotpotQA} & \textbf{2Wiki} & \textbf{Musique} & \textbf{Bamboogle} & \textbf{Average} \\
\midrule
LLM w/o RAG & 0.087 & 0.083 & 0.085 & 0.089 & 0.088 & 0.088 & 0.086 & 0.086 \\
Vanilla RAG & 0.258 & 0.265 & 0.256 & 0.265 & 0.271 & 0.263 & 0.258 & 0.262 \\
RRR \cite{ma2023query} & 1.160 & 1.173 & 1.149 & 1.172 & 1.195 & 1.157 & 1.107 & 1.159 \\
BGM \cite{ke2024bridging} & 0.396 & 0.381 & 0.386 & 0.391 & 0.392 & 0.379 & 0.370 & 0.385 \\
MMOA-RAG \cite{chen2025improving} & 1.709 & 1.673 & 1.690 & 1.688 & 1.676 & 1.669 & 1.594 & 1.671 \\
Self-RAG \cite{asai2023self} & 1.246 & 1.441 & 1.215 & 1.263 & 1.264 & 1.085 & 0.761 & 1.182 \\
Search-o1 \cite{li2025search} & 1.535 & 1.440 & 1.522 & 1.362 & 1.239 & 1.284 & 1.093 & 1.354 \\
\midrule
MAO-ARAG w/o train & 0.162 & 0.244 & 0.246 & 1.124 & 1.163 & 1.049 & 0.674 & 0.666 \\
MAO-ARAG & 0.396 & 0.396 & 0.387 & 1.842 & 1.633 & 1.735 & 1.515 & 1.129 \\
\bottomrule
\end{tabular}
\label{tab:token_costs}
\end{table*}

\begin{table*}[t] \small
\caption{Average number of retrieval call times per query across datasets.}
\centering
\begin{tabular}{>{\centering\arraybackslash}m{4.15cm}
                >{\centering\arraybackslash}m{1.2cm}
                >{\centering\arraybackslash}m{1.2cm}
                >{\centering\arraybackslash}m{1.2cm}
                >{\centering\arraybackslash}m{1.2cm}
                >{\centering\arraybackslash}m{1.2cm}
                >{\centering\arraybackslash}m{1.2cm}
                >{\centering\arraybackslash}m{1.2cm}
                >{\centering\arraybackslash}m{1.4cm}}
\toprule
\textbf{Methods} & \textbf{NQ} & \textbf{PopQA} & \textbf{AmbigQA} & \textbf{HotpotQA} & \textbf{2Wiki} & \textbf{Musique} & \textbf{Bamboogle} & \textbf{Average} \\
\midrule
LLM w/o RAG & 0 & 0 & 0 & 0 & 0 & 0 & 0 & 0 \\
Vanilla RAG & 1.0 & 1.0 & 1.0 & 1.0 & 1.0 & 1.0 & 1.0 & 1.0 \\
RRR \cite{ma2023query} & 4.0 & 4.0 & 4.0 & 4.0 & 4.0 & 4.0 & 3.936 & 3.991 \\
BGM \cite{ke2024bridging} & 1.0 & 1.0 & 1.0 & 1.0 & 1.0 & 1.0 & 1.0 & 1.0 \\
MMOA-RAG \cite{chen2025improving} & 4.0 & 4.0 & 4.0 & 4.0 & 4.0 & 4.0 & 3.936 & 3.991 \\
Self-RAG \cite{asai2023self} & 0.795 & 1.759 & 0.747 & 1.226 & 1.662 & 0.898 & 0.248 & 1.048 \\
Search-o1 \cite{li2025search} & 3.929 & 3.679 & 3.948 & 3.518 & 3.243 & 3.410 & 3.064 & 3.542 \\
\midrule
MAO-ARAG w/o train & 0.410 & 0.854 & 0.862 & 2.712 & 2.802 & 2.534 & 1.864 & 1.720 \\
MAO-ARAG & 1.0 & 1.0 & 1.0 & 3.536 & 3.237 & 3.413 & 3.104 & 2.327 \\
\bottomrule
\end{tabular}
\label{tab:api_times}
\end{table*}

\begin{table*}[t] \small
\caption{Average number of total turns per query across datasets.}
\centering
\begin{tabular}{>{\centering\arraybackslash}m{4.15cm}
                >{\centering\arraybackslash}m{1.2cm}
                >{\centering\arraybackslash}m{1.2cm}
                >{\centering\arraybackslash}m{1.2cm}
                >{\centering\arraybackslash}m{1.2cm}
                >{\centering\arraybackslash}m{1.2cm}
                >{\centering\arraybackslash}m{1.2cm}
                >{\centering\arraybackslash}m{1.2cm}
                >{\centering\arraybackslash}m{1.4cm}}
\toprule
\textbf{Methods} & \textbf{NQ} & \textbf{PopQA} & \textbf{AmbigQA} & \textbf{HotpotQA} & \textbf{2Wiki} & \textbf{Musique} & \textbf{Bamboogle} & \textbf{Average} \\
\midrule
LLM w/o RAG & 1.0 & 1.0 & 1.0 & 1.0 & 1.0 & 1.0 & 1.0 & 1.0 \\
Vanilla RAG & 1.0 & 1.0 & 1.0 & 1.0 & 1.0 & 1.0 & 1.0 & 1.0 \\
RRR \cite{ma2023query} & 2.0 & 2.0 & 2.0 & 2.0 & 2.0 & 2.0 & 2.0 & 2.0 \\
BGM \cite{ke2024bridging} & 1.0 & 1.0 & 1.0 & 1.0 & 1.0 & 1.0 & 1.0 & 1.0 \\
MMOA-RAG \cite{chen2025improving} & 2.0 & 2.0 & 2.0 & 2.0 & 2.0 & 2.0 & 2.0 & 2.0 \\
Self-RAG \cite{asai2023self} & 4.924 & 4.675 & 4.937 & 4.521 & 4.238 & 4.420 & 4.080 & 4.542 \\
Search-o1 \cite{li2025search} & 4.929 & 4.679 & 4.948 & 4.518 & 4.243 & 4.410 & 4.064 & 4.542 \\
\midrule
MAO-ARAG w/o train & 1.033 & 1.167 & 1.090 & 2.997 & 3.159 & 3.014 & 2.128 & 2.084 \\
MAO-ARAG & 1.0 & 1.0 & 1.0 & 4.536 & 4.237 & 4.413 & 4.104 & 2.899 \\
\bottomrule
\end{tabular}
\label{tab:turn_num}
\end{table*}

\begin{table*}[h] %\small
\caption{The prompt of Query Decomposition Serial Agent.}
% \vspace{-2.5mm}
    \centering
    \begin{tcolorbox}[width=1.0\linewidth]
    \textbf{system:} You are a professional assistant skilled at decomposing complex questions into a minimal sequence of logically dependent, independently searchable sub-questions. Each sub-question must: \\
    - Be self-contained and specific \\
    - Be suitable for direct information retrieval from search engines or structured databases \\
    - Be strictly necessary to answer the original question \\
    You must keep the number of sub-questions as low as possible, and never exceed 4 in total. Avoid redundancy and do not include trivial or overly granular sub-questions. \\
    
    \noindent
    \textbf{assistant:} Understood. I will return only factual, retrievable sub-questions, one per line. \\
    
    \noindent
    \textbf{user:} Original question is: \{content of Question\}. \\
    Now decompose the original question into a logically ordered list of sub-questions. \\
    Do not number the sub-questions, write one sub-question per line. \\
    
    \end{tcolorbox}
\label{QDS_prompt}
\end{table*}

\begin{table*}[h] %\small
\caption{The prompt of Query Decomposition Parallel Agent.}
% \vspace{-2.5mm}
    \centering
    \begin{tcolorbox}[width=1.0\linewidth]
    \textbf{system:} You are a professional assistant skilled at decomposing complex multi-entity or multi-location questions into multiple independent and searchable sub-questions. Each sub-question should be specific, logically complete, and not repeat others. \\
    
    \noindent
    \textbf{assistant:} Okay, I will return the parallel sub-questions. \\
    
    \noindent
    \textbf{user:} Original question is \{content of Question\}. \\
    Break down this question into the minimum number of specific, logically complete, and independently searchable sub-questions needed to fully understand and answer the original question. Do not generate more than 4 sub-questions. Each sub-question should be on a separate line, avoid vague demonstratives or repetition, and ensure that each question is self-contained.
    
    \end{tcolorbox}
\label{QDP_prompt}
\end{table*}

\begin{table*}[h] %\small
\caption{The prompt of Query Rewriter Agent.}
% \vspace{-2.5mm}
    \centering
    \begin{tcolorbox}[width=1.0\linewidth]
    \textbf{system:} You are a professional assistant skilled at rewriting overly detailed or redundant questions into a single, concise, and searchable query. Your goal is to keep only the essential part of the question that is needed to find the answer efficiently. \\
    
    \noindent
    \textbf{assistant:} Okay, I will return a concise rewritten query. \\
    
    \noindent
    \textbf{user:} Original question is: \{content of Question\}. \\
    Now rewrite the original question into a single, clear query that focuses only on the essential information needed to find the answer. Avoid unnecessary context, vague references, and maintain specificity. Output only the rewritten query without any extra explanation or formatting. \\
    
    \end{tcolorbox}
\label{QR_prompt}
\end{table*}

\begin{table*}[h] %\small
\caption{The prompt of Document Selector Agent.}
% \vspace{-2.5mm}
    \centering
    \begin{tcolorbox}[width=1.0\linewidth]
    \textbf{system:} You are a helpful, respectful and honest assistant. Your task is to output the ID of the candidate Documents (0, 1, 2,..., n) which are helpful in answering the Question. \\
    
    \noindent
    \textbf{assistant:} Okay, I will provide the ID of candidate Documents which are helpful in answering the Question. \\
    
    \noindent
    \textbf{user:} Question is: \{content of Question\} \\
    \{content of Documents\} \\

    \noindent
    \textbf{assistant:} OK, I received the Question and the candidate Documents.\\

    \noindent
    \textbf{user:} Now, output the ID of the candidate Documents (0,1,2,...,n) which are helpful in answering the Question: \{content of Question\}, for example, in the following format: Document0,Document4,Document6,Document7. \\
    
    \end{tcolorbox}
\label{DS_prompt}
\end{table*}

\begin{table*}[h] %\small
\caption{The prompt of Answer Generator Agent.}
% \vspace{-2.5mm}
    \centering
    \begin{tcolorbox}[width=1.0\linewidth]
    \textbf{system:} You are a helpful, respectful and honest assistant. Your task is to predict the answer to the question based on the given documents. If you don't know the answer to a question, please don't share false information. Answer the question as accurately as possible. \\
    
    \noindent
    \textbf{assistant:} Okay, I will provide the answer to the question based on the corresponding documents. Please provide the question and the corresponding documents. \\
    
    \noindent
    \textbf{user:} Question is: \{content of Question\} \\
    \{content of Documents\} \\
    Now, answer the Question: \{content of Question\}, based on the above Documents.

    \noindent
    \textbf{assistant:} OK, I received the Question and the corresponding Documents. \\

    \noindent
    \textbf{user:} Given the Question and the corresponding Documents, predict the answer to the Question as briefly and accurately as possible based on the Documents. Only give the brief and accurate answer with the form of **answer**. \\
    
    \end{tcolorbox}
\label{AG_prompt}
\end{table*}

\begin{table*}[h] %\small
\caption{The prompt of Answer Summarization Agent.}
% \vspace{-2.5mm}
    \centering
    \begin{tcolorbox}[width=1.0\linewidth]
    \textbf{system:} You are a helpful, respectful and honest assistant. Your task is to predict the final answer to the original question based on the answers to its decomposed sub-questions. If you are not sure about the final answer, do not make up information. Give the most accurate and concise answer possible based on the sub-question answers. \\
    
    \noindent
    \textbf{assistant:} Okay, I will provide the final answer to the original question based on the sub-questions and their corresponding answers. Please provide the original question, the sub-questions, and their answers. \\
    
    \noindent
    \textbf{user:} Original Question:\{content of Question\} \\
    \{content of Context\} \\
    Now, based on the above sub-questions and their answers, answer the Original Question: \{content of Question\}

    \noindent
    \textbf{assistant:} OK, I received the Original Question, its Sub-questions, and their Answers. \\

    \noindent
    \textbf{user:} Given the Original Question, the Sub-questions and their Answers, predict the final answer to the Original Question as briefly and accurately as possible. Only give the brief and accurate answer in the form of **answer**. \\
    
    \end{tcolorbox}
\label{AS_prompt}
\end{table*}

\end{document}